\documentclass[letterpaper]{article} 

\usepackage{aaai25}  
\usepackage{times}  
\usepackage{helvet}  
\usepackage{courier}  
\usepackage[hyphens]{url}  
\usepackage{graphicx} 
\usepackage{natbib}  
\usepackage{caption} 
\usepackage{algpseudocode}  
\usepackage{algorithm}

\usepackage{amssymb} 
\usepackage{amsmath}
\usepackage{amsthm}
\usepackage{xcolor}

\usepackage{newfloat}
\usepackage{listings}
\DeclareCaptionStyle{ruled}{labelfont=normalfont,labelsep=colon,strut=off} 
\floatstyle{ruled}
\newfloat{listing}{tb}{lst}{}
\floatname{listing}{Listing}
\title{
LazyDiT: Lazy Learning for the Acceleration of Diffusion Transformers
}
\author{
    Xuan Shen$^1$\thanks{Work partially done during internship at Adobe Research}, Zhao Song$^2$, Yufa Zhou$^3$, Bo Chen$^4$, Yanyu Li$^1$, Yifan Gong$^1$, Kai Zhang$^2$, Hao Tan$^2$, Jason Kuen$^2$, Henghui Ding$^5$, Zhihao Shu$^6$, Wei Niu$^6$, Pu Zhao$^1$\thanks{Corresponding Author}, Yanzhi Wang$^1$\footnotemark[2], Jiuxiang Gu$^2$\footnotemark[2]
}
\affiliations{
    $^1$Northeastern University \\
    $^2$Adobe Research \\
    $^3$University of Pennsylvania \\ $^4$Middle Tennessee State University \\
    $^5$Fudan University \\
    $^6$University of Georgia

    \{shen.xu, p.zhao, yanz.wang\}@northeastern.edu, wniu@uga.edu,  jigu@adobe.com
}

\theoremstyle{plain}
\newtheorem{theorem}{Theorem} 
\newtheorem{lemma}[theorem]{Lemma}
\newtheorem{definition}[theorem]{Definition}

\newtheorem{fact}[theorem]{Fact}

\newcommand{\wh}{\widehat}
\newcommand{\wt}{\widetilde}

\newcommand{\N}{\mathcal{N}}
\newcommand{\R}{\mathbb{R}}

\renewcommand{\hat}{\wh}

\newcommand{\feed}{\mathrm{feed}}
\newcommand{\attn}{\mathrm{attn}}
\newcommand{\F}{\mathcal{F}}

\DeclareMathOperator*{\E}{{\mathbb{E}}}

\DeclareMathOperator{\diag}{diag}

\DeclareMathOperator{\tr}{tr}

\usepackage{bibentry}

\usepackage{booktabs}
\usepackage{multirow}
\usepackage{multicol}
\usepackage{xcolor}
\usepackage{amsthm}

\newif\ifmodify 
\modifytrue 
\ifmodify
\newcommand{\todo}[1]{\textcolor{red}{#1}}

\else
\newcommand{\todo}[1]{}

\fi

\begin{document}

\maketitle

\begin{abstract}


Diffusion Transformers have emerged as the preeminent models for a wide array of generative tasks, demonstrating superior performance and efficacy across various applications. 
The promising results come at the cost of slow inference, as each denoising step requires running the whole transformer model with a large amount of parameters. 
In this paper, we show that performing the full computation of the model at each diffusion step is unnecessary, as some computations can be skipped by lazily reusing the results of previous steps. 
Furthermore, we show that the lower bound of similarity between outputs at consecutive steps is notably high, and this similarity can be linearly approximated using the inputs. 
To verify our demonstrations, we propose the \textbf{LazyDiT}, a lazy learning framework that efficiently leverages cached results from earlier steps to skip redundant computations. 
Specifically, we incorporate lazy learning layers into the model, effectively trained to maximize laziness, enabling dynamic skipping of redundant computations. 
Experimental results show that LazyDiT outperforms the DDIM sampler across multiple diffusion transformer models at various resolutions.
Furthermore, we implement our method on mobile devices, achieving better performance than DDIM with similar latency.
Code: \textcolor{blue}{\url{https://github.com/shawnricecake/lazydit}}

\end{abstract}

\begin{figure*}[t]
\centering
\includegraphics[width=1.0\linewidth]{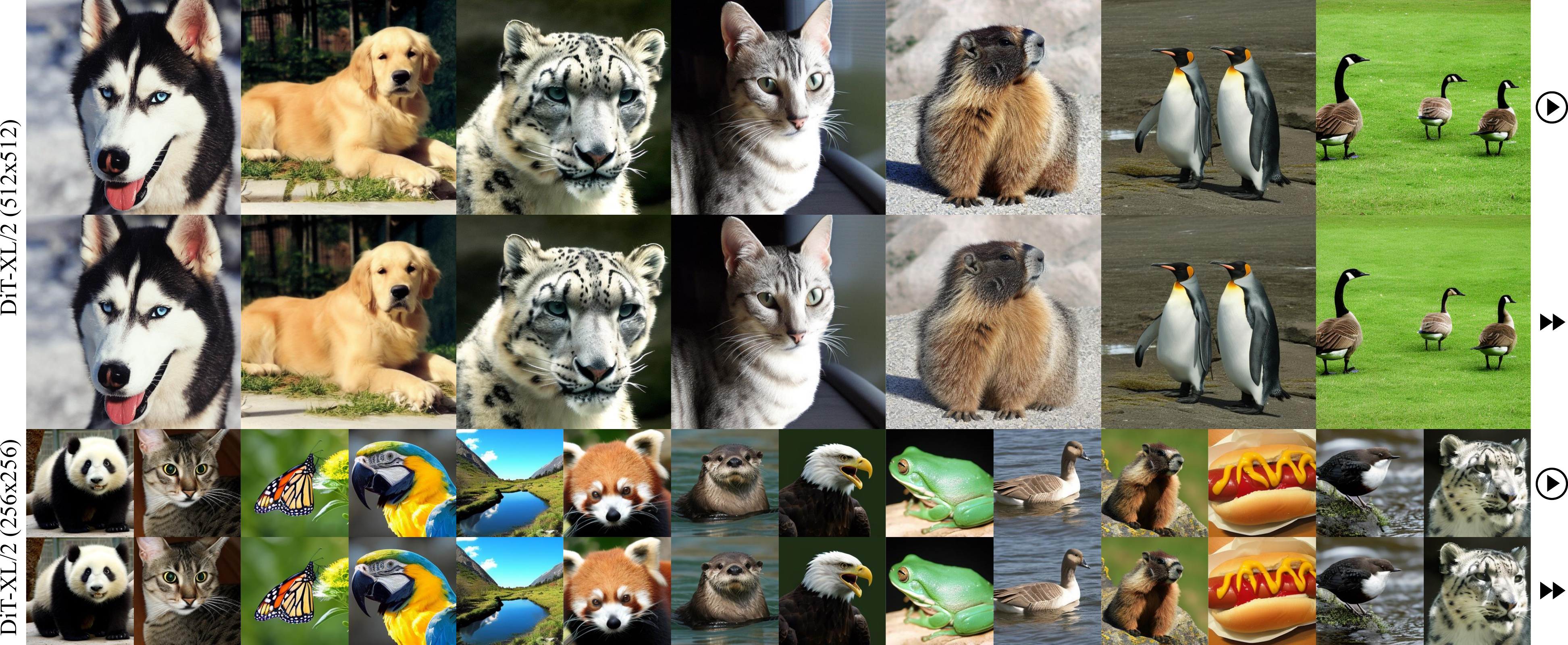}
\caption{
Image generated by DiT-XL/2  in 512$\times$512 and 256$\times$256 resolutions when lazily skipping 50\% computation.
The upper rows display results from original model and the lower rows showcase outcomes of our method.
Our method generates distinct lighting effects for  background and color compared to the baseline, as demonstrated in 
 dog and marmot, respectively.
}
\label{fig:visualization_main}
\end{figure*}

\section{Introduction}

Diffusion models~\cite{ho2020denoising, rombach2022stablediffusion, song2020score, song2019generative, dhariwal2021diffusion,zhan2024fast} have become dominant in image generation research, attributable to their remarkable performance.
U-Net~\cite{ronneberger2015unet} is a widely used backbone in diffusion models, while transformers~\cite{vaswani2017attention} are increasingly proving to be a strong alternative.
Compared to U-Net, transformer-based diffusion models have demonstrated superior performance in high-fidelity image generation~\cite{peebles2023dit, bao2023vitdiffusion}, and their efficacy extends to video generation as well~\cite{lu2023vdt, chen2023gentron, open_sora_plan, opensora}.
This highlights the versatility and potential of transformers in advancing generative tasks across different media.
Despite the notable scalability advantages of transformers, diffusion transformers face major efficiency challenges. 
The high deployment costs and the slow inference speeds create the significant barriers to their practical applications \cite{zhan2024exploring,zhan-etal-2024-rethinking-token,zhan2021achieving,wu2022compiler,li2022pruning,yang2023pruning}, which motivates us to explore their acceleration methods.

The increased sampling cost in diffusion models stems from two main components: the numerous timesteps required and the computational expense associated with each inference step.
To improve sampling efficiency, existing methods generally fall into two categories: reducing the total number of sampling steps~\cite{song2020ddim, liu2022pseudo, bao2022analytic,zhan2024fast} or lowering the computational cost per step~\cite{yang2023diffusion, he2023ptqd}.
Several works~\cite{yin2024onestep, luo2024latent, salimans2022progressive} employ distillation techniques to reduce the number of sampling steps. 
Conversely, works~\cite{li2023snapfusion, kim2023bksdm, fang2023structural, li2023qdiffusion} utilize compression techniques to streamline diffusion models. 
Recently, some studies have introduced caching mechanisms into the denoising process~\cite{ma2023deepcache, wimbauer2023cache} to accelerate the sampling. 
However, previous compression approaches of diffusion models 
have primarily focused on optimizing  U-Net, leaving transformer-based  models largely unexplored.

Leveraging characteristic of uniquely structured, prior compression works~\cite{zhang2024laptopdifflayerpruningnormalized, raposo2024mixtureofdepthsdynamicallyallocatingcompute, fan2019reducingtransformerdepthdemand,kong2022spvit,kong2023peeling,zhang2022advancing,rtseg,zhao-etal-2024-pruning,shen2024search,shen2024agile,shen2024edgeqat,shen2024hotaq,shen2023deepmad} have concentrated on techniques such as layer pruning and width pruning.
However, we observe that removing certain layers results in a significant performance drop. 
This indicates the redundancy in diffusion transformers primarily occurs between sampling steps rather than the model architecture. This finding forms basis for exploring methods to reduce  frequency of layer usage, aiming to decrease computational costs and accelerate the  diffusion.

In this paper, we propose \textbf{LazyDiT}, a cache-based approach designed to dynamically reduce computational costs and accelerate the diffusion process.
We begin by analyzing the output similarity between the current and previous steps, identifying that the lower bound of this similarity is notably high during the diffusion process.
Then, we delve deeper into the similarity using a Taylor expansion around the current input, revealing that the similarity can be linearly approximated.
Building on the theoretical analysis, we implement a lazy learning framework by introducing linear layers before each Multi-Head Self-Attention (MHSA) and pointwise feedforward (Feedforward) module.
These added layers are trained with the proposed lazy loss to learn whether the subsequent module can be lazily bypassed by leveraging the previous step's cache.
Compared to the DDIM sampler, extensive experiments demonstrate that our method achieves superior performance with similar computational costs.
As shown in Figure~\ref{fig:visualization_main}, by lazily skipping 50\% of the computations, our method achieves nearly the same performance as the original diffusion process.
We also profile the latency of the diffusion process on mobile devices to offer a detailed comparison with the DDIM sampler.
Our results show our superior image generation quality than DDIM with similar latency. 
Our main contributions  are summarized as follows,
\begin{itemize}
    \item We explore the redundancy in diffusion process by evaluating the similarity between module outputs at consecutive steps, finding that the lower bound of the similarity is notably high.

    \item We establish that the lazy skip strategy can be effectively learned through a linear layer based on the Taylor expansion of similarity.
    
    \item We propose a lazy learning framework to optimize the diffusion process in transformer-based models by lazily bypassing computations using the previous step's cache.
 
    \item Experiments show that the proposed method achieves better performance than DDIM sampler. We further implement our method on mobile devices, showing that our method is a promising solution for real-time generation.
    
\end{itemize}

\section{Related Work}

\paragraph{Transformer-Based Diffusion Models.}
Recent works such as GenVit~\cite{yang2022genvit}, U-Vit~\cite{bao2023vitdiffusion}, DiT~\cite{peebles2023dit}, LlamaGen~\cite{sun2024autoregressive}, and MAR~\cite{li2024autoregressive} have incorporated transformers~\cite{vaswani2017attention} into diffusion models, offering a different approach compared to the traditional U-Net architecture. GenViT incorporates the ViT~\cite{dosovitskiy2021imageworth16x16words,yize2024neural,yize2024pruning} architecture into DDPM, while U-ViT further enhances this approach by introducing long skip connections between shallow and deep layers. DiT demonstrates the scalability of diffusion transformers, and its architecture has been further utilized for text-to-video generation tasks, as explored in works~\cite{sora}. LlamaGen introduces autoregressive models to image generation, verifying the effectiveness of the 'next-token prediction' in this domain. 
Thus, it is crucial to explore efficient designs for those large models to accelerate the diffusion process.

\paragraph{Acceleration for Diffusion Models.}
High-quality image generation with diffusion models necessitates multiple sampling steps, leading to increased latency~\cite{gong20242,shen2022data}.
To enhance efficiency, DDIM~\cite{song2020ddim} extends original DDPM to non-Markovian cases when DPM-Solver~\cite{lu2022dpmsolverfastodesolver} advances the approximation of diffusion ODE solutions.
Regarding the works that require fine-tuning, such as~\cite{lin2024sdxllightningprogressiveadversarialdiffusion, yin2024onestep}, they employ distillation techniques to effectively reduce the number of sampling steps.
Additionally, reducing the computational workload for each diffusion step is a widely adopted, strategy to enhance the efficiency of the diffusion process. Various approaches have been explored, such as works~\cite{fang2023structural, castells2024ldprunerefficientpruninglatent, wang2024sparsedmsparseefficientdiffusion, zhang2024laptopdifflayerpruningnormalized} that adopt weight pruning techniques, works~\cite{he2023ptqd, li2023qdiffusion} that employ quantization techniques, and even works~\cite{kim2023bksdm, li2023snapfusion} that redesign the architecture of diffusion models.  



\section{Methodology}

\subsection{Preliminaries}
\paragraph{Notations.}
We use $\E[]$ to denote the expectation. 
We use ${\bf 1}_n$ to denote a length-$n$ vector where all the entries are ones. 
We use $x_{i,j}$ to denote the $j$-th coordinate of $x_i \in \mathbb{R}^n$.
We use $\|x\|_p$ to denote the $\ell_p$ norm of a vector $x$. 
We use $\|A\|$ to denote the spectral norm for a matrix $A$. 
We use $a \circ b$ to denote the element-wise product of two vectors $a, b$.
For a tensor $X \in \R^{B \times N \times D}$ and a matrix $U \in \R^{D \times d_1}$, we define $Y = X \cdot U \in \R^{B \times N \times d_1}$.
For a matrix $V \in \R^{d_2 \times B}$ and a tensor $X \in \R^{B \times N \times D}$, we define $Z =  V \cdot X \in \R^{d_2 \times N \times D}$.
For a square matrix $A$, we use $\tr[A]$ to denote the trace of $A$.
For two matrices $X, Y$, the standard inner product between matrices is defined by $\langle X, Y \rangle := \tr[X^\top Y]$. 
We use $U(a,b)$ to denote a uniform distribution.
We use $\N(\mu, \sigma^2)$  to denote a Gaussian distribution.
We define cosine similarity as $f(X,Y) = \frac{\tr[X^\top Y]}{\|X\|_F \cdot \|Y\|_F}$ for matrices $X,Y$.

\paragraph{Diffusion Formulation.}
Diffusion models~\cite{ho2020denoising, song2020score} operate by transforming a sample $x$ from its initial state within a real data distribution $p_{\beta}(x)$ into a noisier version through diffusion steps.
For a diffusion model $\epsilon_{\theta}(\cdot)$ with parameters $\theta$, the training objective~\cite{sohldickstein2015deepunsupervisedlearningusing} can be expressed as follows,
\begin{align*}
    \min_{\theta} \E_{t \sim U[0,1], x \sim p_{\beta}(x), \epsilon \sim \mathcal{N}(0, I)  }  \|  \epsilon_{\theta}( t, z_t  ) - \epsilon  \|_2 , 
\end{align*}
where $t$ denotes the timestep; $\epsilon$ denotes the ground-truth noise; $z_t = \alpha_t \cdot x + \sigma_t \cdot \epsilon$ denotes the noisy data; $\alpha_t$ and $\sigma_t$ are the strengths of signal and noise. 

For comparison purposes, this paper adopts Denoising Diffusion Implicit Models (DDIM)~\cite{song2020ddim} as sampler. 
The iterative denoising process from timestep $t$ to the previous timestep $t'$ is described as follows,
\begin{align*}
    z_{t'} = \alpha_{t'} \cdot \frac{z_t - \sigma_t \epsilon_{\theta}(t, z_t)}{\alpha_t} + \sigma_{t'} \cdot \epsilon_{\theta} (t, z_t) ,
\end{align*}
where $z_{t'}$ is iteratively fed to $\epsilon_{\theta}(\cdot)$ until $t'$ becomes $0$.

\paragraph{Latent Diffusion Models.}
The Latent Diffusion Model (LDM)~\cite{rombach2022highresolutionimagesynthesislatent} decreases computational demands and the number of steps with the latent space, which is obtained by encoding with a pre-trained variational autoencoder (VAE)~\cite{sohldickstein2015deepunsupervisedlearningusing}. 
Besides, the classifier-free guidance (CFG)~\cite{ho2022classifierfreediffusionguidance} is adopted to improve quality as follows,
\begin{align*}
    \hat{\epsilon}_{\theta}(t, z_t, c) = w \cdot \epsilon_{\theta}(t, z_t, c) - (w-1) \cdot \epsilon_{\theta}(t, z_t, c_\phi),
\end{align*}
where $\epsilon_{\theta}(t, z_t, c_\phi)$ denotes unconditional prediction with null text; $w$ denotes guidance scale which is used as control of conditional information and $w \geq 1$. 

\subsection{Similarity Establishment}
Let $B$ be the number of batches, $N$ be the number of patches, $D$ be the hidden dimension, $T$ be the number of diffusion steps, and $L$ be the number of model layers in diffusion transformers.
Let $f(\cdot, \cdot): \R^{B \times N \times D} \times \R^{B \times N \times D} \to [0,1]$ be the function that estimate the similarity between two variables. 
Let $\mathcal{F}^{\Phi}_l(\cdot): \R^{B \times N \times D} \to \R^{B \times N \times D}$ be the Multi-Head Self-Attention (MHSA) / pointwise feedforward (Feedforward) module at $l$-th layer where $\Phi \in \{ \attn, \feed \}$ for $l \in [L]$.
We use normalized $X_{l,t}^{\Phi} \in \R^{B \times N \times D}$ to denote the input hidden states with scaling factor $a_t$ and shifting factor $b_t$ at timestep $t$ in the $l$-th layer for $t \in [T], l \in [L]$.
We denote the output at $l$-th layer and $t$-th timestep as 
$Y^\Phi_{l,t}$.

\paragraph{Impact of Scaling and Shifting.}

As progressive denoising proceeds, the input difference between, $X_{l,t-1}^{\Phi}$ and $X_{l,t}^{\Phi}$, grows. In contrast, the application of scaling and shifting transformations introduces an alternate problem formulation, potentially affecting the input distance in a manner that requires a different analytical approach.
In detail, the diffusion transformer architecture incorporates scaling and shifting mechanisms in the computation of both the MHSA and Feedforward modules, utilizing the embeddings of the timestep, $\mathrm{emd}(t) \in \mathbb{R}^D$, and the condition, $\mathrm{emd}(c) \in \mathbb{R}^D$. 
We define $y_t = \mathrm{SiLU}(\mathrm{emd}(t) + \mathrm{emd}(c)) \in \mathbb{R}^D$, with corresponding scaling and shifting factors defined as follows, 
\begin{itemize} 
    \item Scaling factor: $a_t = W_{l,a} \cdot y_t + v_{l,a} \in \R^D$;
     \item Shifting factor: $b_t = W_{l,b} \cdot y_t + v_{l,b} \in \R^D$;
\end{itemize} 
where $W_{l,a},W_{l,b} \in \R^{D\times D}, v_{l,a},v_{l,b}\in \R^D$ are the linear projection weight and bias, respectively.

Meanwhile, we define broadcasted matrices to represent the scaling and shifting factors, ensuring the alignment with the implementation of diffusion transformers as follows,
\begin{itemize}
\item Let $A_t \in \R^{N \times D}$ be defined as the matrix that all rows are $a_t$, i.e. $(A_t)_i := a_t$ for $i \in [N]$, and $A_{t-1}, B_t, B_{t-1}$ can be defined as the same way. 
\end{itemize}

Then, we deliver the demonstration showing that there exist $y_t,y_{t-1}$ such that, after scaling and shifting, the distance between inputs $X_{l,t-1}^{\Phi}, X_{l,t}^{\Phi}$, defined in the Left Hand Side of the following Eq.~\eqref{eq:A_t_X_t_plus_B_t}, can be constrained within a small bound.  
Given $a_t$ and $b_t$ are both linear transformation of $y_t$, the problem reduces to demonstrating the existence of vectors $a_t$, $b_t$, $a_{t-1}$, and $b_{t-1}$ that satisfy following conditions, 
\begin{align}\label{eq:A_t_X_t_plus_B_t}
    \| (A_{t-1} \circ X_{l,t-1}^{\Phi} + B_{t-1}) - (A_{t} \circ X_{l,t}^{\Phi} + B_{t}) \| \leq \eta ,
\end{align}
where $\eta \in (0, 0.1)$. 
And Eq.~\eqref{eq:A_t_X_t_plus_B_t} is equivalent as follows,
\begin{align}\label{eq:A_B_C_eta}
    \| A  \circ X_{l,t-1}^{\Phi} + B \circ X_{l,t}^{\Phi} + C \|_F \leq \eta  ,
\end{align} 
where $A := A_{t-1}, B := - A_t, C:= B_{t-1} - B_t$. 

We identify that there exists $a$, $b$ and $c$ such that Eq.~\eqref{eq:A_B_C_eta} holds, and the detailed demonstration and explanation are included in Lemma 12 at Appendix C.2.
Subsequently, we generate the following theorem, 
\begin{theorem}[Scaling and shifting, informal version of Theorem 13 at Appendix C.2]\label{thm:scaling_shifting:informal}
There exist time-variant and condition-variant scalings and shiftings such that the distance between two inputs at consecutive steps for MHSA or Feedforward is bounded.
\end{theorem}



\paragraph{Similarity Lower Bound.}
To leverage the cache from the previous step, we begin by investigating the cache mechanism in transformer-based diffusion models.
This analysis focuses on the similarity between the current output and the preceding one, providing insights into the efficacy of reusing cached information. 
One typical transformer block consists of two primary modules: MHSA module and Feedforward module.
Both modules are computationally expensive, making them significant contributors to the overall processing cost. 
Thus, we aim to examine the output similarities between the current and previous steps for both modules. 
By identifying cases of high similarity, we can skip redundant computations, thereby reducing the overall computational cost. 
In practice, we employ the cosine similarity $f(\cdot, \cdot)$ for the computation of similarity as follows, 
\begin{align}\label{eq:cosine_similarity}
    f(Y^\Phi_{l,t-1}, Y^\Phi_{l,t})  = \frac{\tr[(Y^\Phi_{l,t-1})^\top \cdot Y^\Phi_{l,t} ]}{\|Y^\Phi_{l,t-1}\|_F \cdot \|Y^\Phi_{l,t}\|_F}.
\end{align}

Inspired by the Lipschitz property~\cite{t13}, we transform the similarity measure into a distance metric, defined as \(\mathrm{Dist} := \|  Y^\Phi_{l,t-1} - Y^\Phi_{l,t} \|\), to simplify the analysis of similarity variations. 
According to Fact 7 in Appendix B.2, 
similarity function $f(\cdot,\cdot)$ is further transformed as follows,
\begin{align*}
    f(Y^\Phi_{l,t-1}, Y^\Phi_{l,t}) = 1 - \mathrm{Dist}/2.
\end{align*}

For convenience, we further define the hidden states after scaling and shifting as $Z_{l,t}^{\Phi} := A_t \circ X_{l,t}^{\Phi} + B_t$. 
Meanwhile, building upon the Lemma H.5 of~\cite{dsxy23}, we further derive the upper bounds for the distance \(\mathrm{Dist}\) for either the MHSA or Feedforward modules as follows,
\begin{align}
\label{eq:dist}
        \mathrm{Dist} \leq C \cdot \|  Z_{l,t-1}^{\Phi} - Z_{l,t}^{\Phi} \| .
\end{align}
where $C$ is the Lipschitz constant related to the module. 

Subsequently, with Theorem~\ref{thm:scaling_shifting:informal}, we integrate Eq.~\eqref{eq:A_t_X_t_plus_B_t} and derive the bound of the similarity as follows, 
\begin{align*}
    f(Y^\Phi_{l,t-1}, Y^\Phi_{l,t}) \geq & ~ 1- \alpha ,
\end{align*}
for $\alpha := O(C^2 \eta^2)$ and $\eta$ is sufficiently small in practice. 


Thus, we deliver Theorem~\ref{thm:similar_or_dissimilar:informal} as below, which asserts that the lower bound of the output similarity between the two consecutive sampling steps is high. 
\begin{theorem}[Similarity lower bound, informal version of Theorem 18 at Appendix C.4]\label{thm:similar_or_dissimilar:informal}
The lower bound of the similarity $f(Y^\Phi_{l,t-1}, Y^\Phi_{l,t})$ between the outputs at timestep $t-1$ and timestep $t$ is high.
\end{theorem}

\paragraph{Linear Layer Approximation.}
The similarity can be approximated using the inputs from either the current step $Z_{l,t}^{\Phi}$ or previous one $Z_{l,t-1}^{\Phi}$, due to its mathematical symmetry according to Eq.~\eqref{eq:cosine_similarity}.
We then apply the Taylor expansion around $Z_{l,t}^{\Phi}$ as follows,
\begin{align*}
    & ~ f(Y^\Phi_{l,t-1}, Y^\Phi_{l,t}) \\
    = & ~  \tr[(Y^\Phi_{l,t-1})^\top \cdot (0 + J \cdot Z_{l,t}^{\Phi} + O(1))],
\end{align*} 
where $J$ is the Jacobian matrix.  

Through Taylor expansion, we identify that there exists a $W_l\in \R^{D \times D_{\mathrm{out}}}$ along with $Z_{l,t}^{\Phi}$ such that the similarity can be linearly approximated with certain error as follows,
\begin{align*}
   \langle W^\Phi_l, Z_{l,t}^{\Phi} \rangle = f(Y^\Phi_{l,t-1}, Y^\Phi_{l,t}) + O(1),
\end{align*}
where the detailed proof is included in Appendix C.5 Eq.(9).

Then, we generate the Theorem~\ref{thm:linear_layer:informal} as follows,
\begin{theorem}[Linear layer approximation, informal version of Theorem 19 at Appendix C.5]\label{thm:linear_layer:informal}
The similarity function $f(\cdot, \cdot)$ can be approximated by a linear layer with respect to the current input, i.e. $f(Y^\Phi_{l,t-1}, Y^\Phi_{l,t}) = \langle W^\Phi_l, Z_{l,t}^{\Phi} \rangle$ where $W^\Phi_l$ is the weight of a linear layer for MHSA or Feedforward in the $l$-th layer of diffusion model.
\end{theorem}
In our experiments, we utilize the lazy learning framework to obtain $ W^\Phi_l $, and we set $ D_\mathrm{out} = 1 $ to minimize computational cost.
The details of this approach are explained in the following section.




\begin{figure}[h]
  \centering
  \includegraphics[width=1.0\linewidth]{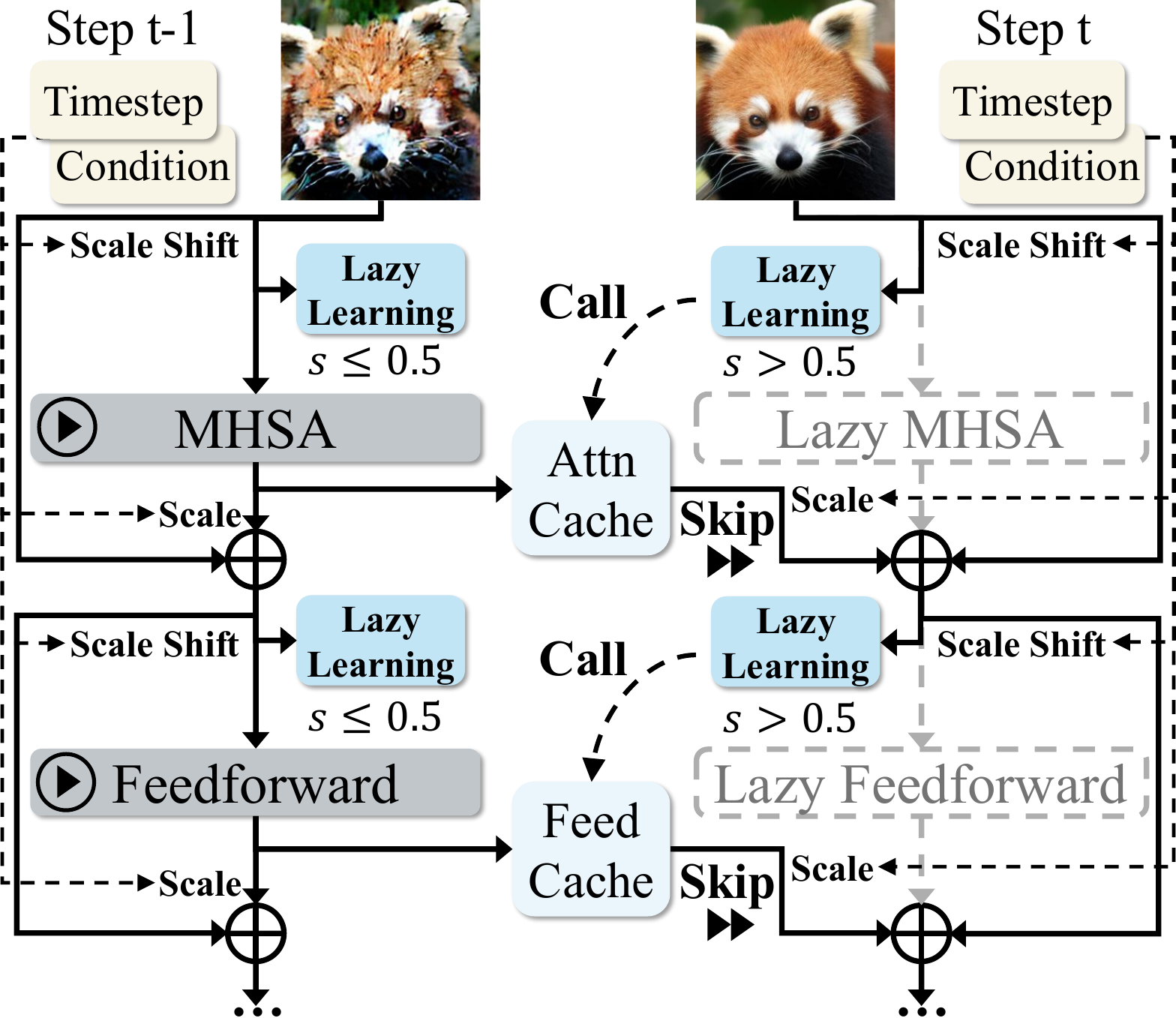}
  \caption{
  Overview framework. We skip the computation of MHSA or Feedforward by calling the previous step cache.
  }
  \label{fig:skip_overview}
\end{figure}

\subsection{Lazy Learning}

As illustrated in Figure~\ref{fig:skip_overview}, we incorporate lazy learning linear layers before each MHSA module and Feedforward module to learn the similarity.
The MHSA module or Feedforward module is bypassed and replaced with the cached output from the previous step if the learned similarity is below 0.5.
The input scale, input shift, output scale, and residual connections remain unchanged from the normal computation.
The training details and the calculation of lazy ratio are outlined in the following paragraphs.

\paragraph{Training Forward.}


Assume we add linear layers with weights $W_{l}^{\Phi} \in \R^{D \times 1}$ for each module $\mathcal{F}_l^{\Phi}(\cdot)$ at $l$-th layer in the model.
For input hidden states $X_{l,t}^{\Phi} \in \R^{B \times N \times D}$ for the module at $l$-th layer and $t$-th step, the similarity $s_{l,t}^{\Phi} \in \R^{B}$ of the module is computed as follows,
\begin{align*}
    s_{l,t}^{\Phi} = \text{sigmoid} ( ( Z_{l,t}^{\Phi} \cdot  W_{l}^{\Phi} )\cdot {\bf 1}_N  ) .
\end{align*}



We then define the forward pass of the MHSA module or Feedforward module at $l$-th layer and $t$-th step with the input $X_{l,t}^{\Phi}$ during the training progress as follows,
\begin{align*}
    Y_{l,t}^{\Phi} = & ~ \diag( {\bf 1}_B - s_{l,t}^{\Phi} )  \cdot \mathcal{F}_{l}^{\Phi} (Z_{l,t}^{\Phi}) \\
    & ~ +  \diag(  s_{l,t}^{\Phi} ) \cdot  Y_{l,t-1}^{\Phi} .
\end{align*}

\paragraph{Backward Loss.}
Alongside the diffusion loss for a given timestep $t$ during training, we introduce a lazy loss to encourage the model to be more lazy—relying more on cached computations rather than diligently executing the MHSA modules or Feedforward modules, as follows,
\begin{align}\label{eq:penalty_ratio}
    \mathcal{L}_{t}^{\mathrm{lazy}} = & ~ \rho^{\attn} \cdot \frac{1}{B} \sum_{l=1}^{L} \sum_{b=1}^{B}  (1-(s_{l,t}^{\attn})_b) \notag \\
    & + ~ \rho^{\feed} \cdot \frac{1}{B} \sum_{l=1}^{L} \sum_{b=1}^{B} (1-(s_{l,t}^{\feed})_b),
\end{align}
where the
$\rho^{\attn}$ and $\rho^{\feed}$ denote the penalty ratio of MHSA module and Feedforward module, respectively.

We combine the lazy loss with diffusion loss and regulate $\rho^{\attn}$ and $\rho^{\feed}$ to control the laziness (i.e., number of skips with cache) of sampling with diffusion transformers.

\paragraph{Accelerate Sampling.}
After finishing the lazy learning with a few steps, we then accelerate the sampling during the diffusion process as follows,
\begin{align*}
    Y_{l,t}^{\Phi} =  
    \left\{
    \begin{array}{ll}
    \mathcal{F}_{l}^{\Phi} (Z_{l,t}^{\Phi}), & \ s_{l,t}^{\Phi} \leq 0.5, \\
    Y_{l,t-1}^{\Phi}, & \ s_{l,t}^{\Phi} > 0.5,
    \end{array}
    \right.
\end{align*}
where $\Phi \in \{ \attn, \feed \}$ can be either MHSA module or Feedforward module, and the skip occurs when $s_{l,t}^{\Phi} > 0.5$.

Then, the lazy ratio $\Gamma^{\Phi} \in \mathbb{Z}^{B}$ of MSHA or Feedforward for $B$ batches during sampling is be computed as follows,
\begin{align*}
    \Gamma^{\Phi} = \frac{1}{LT} \sum_{l=1}^{L} \sum_{t=1}^{T} \left \lceil  s^{\Phi}_{l,t} - 0.5 \right \rceil.
\end{align*}



\section{Experimental Results}

\subsection{Experiment Setup}

\paragraph{Model Family.}
We validate the effectiveness of our method on both the DiT~\cite{peebles2023dit} and LargeDiT~\cite{largedit} model families.
Specifically, our experiments utilize the officially provided models including DiT-XL/2 (256$\times$256), DiT-XL/2 (512$\times$512), Large-DiT-3B (256$\times$256), and Large-DiT-7B (256$\times$256).

\paragraph{Lazy Learning.}
We freeze the original model weights and introduce linear layers as lazy learning layers before each MHSA and Feedforward module at every diffusion step.
For various sampling steps, these added layers are trained on the ImageNet dataset with 500 steps, with a learning rate of 1e-4 and using the AdamW optimizer.
Following the training pipeline in DiT, we randomly drop some labels, assign a null token for classifier-free guidance, and set a global batch size of 256.
The training is conducted on 8$\times$NVIDIA A100 GPUs within 10 minutes.

\paragraph{Penalty Regulation.}
We regulate the penalty ratios $\rho^{\attn}$ and $\rho^{\feed}$ for MHSA and Feedforward in Eq.~\eqref{eq:penalty_ratio} from 1e-7 to 1e-2.
Both penalty ratios are kept identical in our experiments to optimize performance, as explained by the ablation study results shown in the lower of Figure~\ref{fig:ablation_two_figure}.

\begin{table}[t] 
\centering
\resizebox{1.0\linewidth}{!}{
\begin{tabular}{c|c|c|ccccc}
\toprule
\multicolumn{1}{c|}{\multirow{2}{*}{Method}} & \multicolumn{1}{c|}{\# of} & Lazy  & \multicolumn{1}{c}{FID} & \multicolumn{1}{c}{sFID} & \multicolumn{1}{c}{IS} & \multicolumn{1}{c}{Prec.} & \multicolumn{1}{c}{Rec.} \\
\multicolumn{1}{c|}{} & Step & Ratio & $\downarrow$ & $\downarrow$ & $\uparrow$ & $\uparrow$ & $\uparrow$ \\
\midrule
\multicolumn{8}{c}{DiT-XL/2 (256$\times$256)}                                                        \\
\midrule
DDIM   & 50        & /  & 2.34           & 4.33           & 241.01          & 80.13          & 59.55          \\
\midrule
DDIM   & 40        & /  & 2.39           & 4.28           & 236.26          & 80.10          & 59.41          \\
Ours   & 50        & 20\%  & \textbf{2.37}  & 4.33  & \textbf{239.99} & \textbf{80.19} & \textbf{59.63} \\
\midrule
DDIM   & 30        & /  & 2.66           & 4.40           & 234.74          & 79.85          & 58.96          \\
Ours   & 50        & 40\%  & \textbf{2.63}  & \textbf{4.35}  & \textbf{235.69} & 79.59          & 58.94          \\
\midrule
DDIM   & 25        & /  & 2.95           & 4.50           & 230.95          & 79.49          & 58.44          \\
Ours   & 50        & 50\%  & \textbf{2.70}  & \textbf{4.47}  & \textbf{237.03} & \textbf{79.77} & \textbf{58.65} \\
\midrule
DDIM   & 20        & /  & 3.53           & 4.91           & 222.87          & 78.43          & 57.12          \\
Ours   & 40        & 50\%  & \textbf{2.95}  & \textbf{4.78}  & \textbf{234.10} & \textbf{79.61} & \textbf{57.99} \\
\midrule
DDIM   & 14        & /  & 5.74           & 6.65           & 200.40          & 74.81          & 55.51          \\
Ours   & 20        & 30\%  & \textbf{4.44}  & \textbf{5.57}  & \textbf{212.13} & \textbf{77.11} & \textbf{56.76} \\
\midrule
DDIM   & 10        & /  & 12.05          & 11.26          & 160.73          & 66.90          & 51.52          \\
Ours   & 20        & 50\%  & \textbf{6.75}  & \textbf{8.53}  & \textbf{192.39} & \textbf{74.35} & \textbf{52.43} \\
\midrule
DDIM   & 7         & /  & 34.14 &	27.51 &	91.67 &	47.59 &	46.83 \\
Ours   & 10        & 30\%  & \textbf{17.05} & \textbf{13.37} & \textbf{136.81} & \textbf{62.07} & \textbf{50.37} \\
\midrule
\multicolumn{8}{c}{DiT-XL/2 (512$\times$512)}                                                                                                \\
\midrule
DDIM   & 50        & / & 3.33           & 5.31           & 205.01          & 80.59          & 55.89          \\
\midrule
DDIM   & 30        & / & 3.95           & 5.71           & 195.84          & 80.19          & 54.52          \\
Ours   & 50        & 40\% & \textbf{3.67}  & \textbf{5.65}  & \textbf{202.25} & 79.80 & \textbf{55.17} \\
\midrule
DDIM   & 25        & / & 4.26           & 6.00           & 192.71          & 79.37          & 53.99          \\
Ours   & 50        & 50\% & \textbf{3.94}  & \textbf{5.92}  & \textbf{200.93} & \textbf{80.47} & \textbf{54.05} \\
\midrule
DDIM   & 20        & /  & 5.12           & 6.60           & 184.23          & 78.37          & 53.50          \\
Ours   & 40        & 50\%  & \textbf{4.32}  & \textbf{6.50} & \textbf{196.01} & \textbf{79.64} & 53.19  \\
\midrule
DDIM   & 10        & /  & 14.76          & 12.38          & 129.19          & 65.51          & 48.95          \\
Ours   & 20        & 50\%  & \textbf{9.18}  & \textbf{10.85} & \textbf{160.30} & \textbf{73.16} & \textbf{49.27}  \\
\midrule
DDIM   & 8         & /  & 24.22          & 18.89          & 100.75          & 55.49          & 46.93          \\
Ours   & 10        & 20\%  & \textbf{16.28}  & \textbf{11.44} & \textbf{123.65} & \textbf{64.36} & \textbf{49.99}  \\
\bottomrule
\end{tabular}
}
\caption{
DiT model results on ImageNet (cfg=1.5).
`Lazy Ratio' indicates the percentage of skipped MHSA and Feedforward modules during diffusion process. 
}
\label{tab:main_results}
\end{table}

\paragraph{Evaluation.}
To evaluate the effectiveness of our method, we primarily compare our method to the DDIM~\cite{song2020ddim}, varying the sampling steps from 10 to 50.
Visualization results are generated with DiT-XL/2 model in 256$\times$256 and 512$\times$512 resolutions.
For quantitative analysis, the 50,000 images are generated per trial with classifier-free guidance in our experiments.
We adopt the Fréchet inception distance (FID)~\cite{heusel2017fid}, Inception Score (IS)~\cite{salimans2016is}, sFID~\cite{nash2021sfid}, and Precision/Recall~\cite{Kynkaanniemi2019precisionrecall} as the evaluation metrics.
The computation cost as TMACs is calculated with the work~\cite{opcounter}.

\paragraph{Testing Bed.}
We implement our acceleration framework on mobile devices, specifically, we use OpenCL for mobile GPU backend. LazyDiT is built upon our existing DNN execution framework
that supports extensive operator fusion for various DNN structures. We also integrated other general DNN inference optimization methods similar to those in ~\cite{tvm, tflite}, including memory layout and computation graph.
Results are obtained using a smartphone with a Qualcomm Snapdragon 8 Gen 3, featuring a Qualcomm Kryo octa-core CPU, a Qualcomm Adreno GPU, and 16 GB of unified memory.
Each result take 50 runs, with average results reported as variance is negligible.


\begin{table}[t] 
\centering
\resizebox{1.0\linewidth}{!}{
\begin{tabular}{c|c|c|ccccc}
\toprule
\multicolumn{1}{c|}{\multirow{2}{*}{Method}} & \multicolumn{1}{c|}{\# of} & Lazy  & \multicolumn{1}{c}{FID} & \multicolumn{1}{c}{sFID} & \multicolumn{1}{c}{IS} & \multicolumn{1}{c}{Prec.} & \multicolumn{1}{c}{Rec.} \\
\multicolumn{1}{c|}{} & Step & Ratio & $\downarrow$ & $\downarrow$ & $\uparrow$ & $\uparrow$ & $\uparrow$ \\
\midrule
\multicolumn{7}{c}{Large-DiT-3B (256$\times$256)}                                                                          \\
\midrule
DDIM   & 50 & /       & 2.10           & 4.36           & 263.83          & 80.36          & 59.55          \\
\midrule
DDIM   & 35 & /       & 2.23           & 4.48           & 262.22          & 80.08          & 60.33          \\
Ours   & 50 & 30\%       & \textbf{2.12}  & \textbf{3.32}  & \textbf{262.27} & \textbf{80.27} & 60.01          \\
\midrule
DDIM   & 25 & /       & 2.75           & 4.95           & 247.68          & 79.12          & 58.88          \\
Ours   & 50 & 50\%       & \textbf{2.42}  & \textbf{4.86}  & \textbf{257.59} & \textbf{79.71} & \textbf{59.41} \\
\midrule
DDIM   & 20 & /       & 3.46           & 5.57           & 239.18          & 77.87          & 58.71          \\
Ours   & 40 & 50\%       & \textbf{2.79}  & \textbf{5.15}  & \textbf{250.84} & \textbf{78.84} & \textbf{59.42} \\
\midrule
DDIM   & 14 & /       & 5.84           & 7.80           & 211.13          & 73.96          & 56.32          \\
Ours   & 20 & 30\%       & \textbf{4.64}  & \textbf{6.35}  & \textbf{220.48} & \textbf{75.61} & \textbf{57.81} \\
\midrule
DDIM   & 10 & /       & 13.05          & 14.17          & 162.22          & 64.89          & 51.92          \\
Ours   & 20 & 50\%       & \textbf{7.36}  & \textbf{10.55} & \textbf{197.67} & \textbf{72.14} & \textbf{54.40} \\
\midrule
DDIM   & 7  & /       & 37.33          & 35.02          & 84.68           & 44.52          & 47.17          \\
Ours   & 10 & 30\%       & \textbf{16.40} & \textbf{12.72} & \textbf{143.70} & \textbf{61.31} & \textbf{54.17} \\
\midrule
\multicolumn{7}{c}{Large-DiT-7B (256$\times$256)}                                                                          \\
\midrule
DDIM   & 50 & /       & 2.16           & 4.64           & 274.89          & 80.87          & 60.00          \\
\midrule
DDIM   & 35 & /       & 2.29           & 4.83           & 267.31          & 80.42          & 59.21          \\
Ours   & 50 & 30\%       & \textbf{2.13}  & \textbf{4.49}  & \textbf{267.37} & \textbf{80.55} & \textbf{60.76} \\
\midrule
DDIM   & 25 & /       & 2.76           & 5.36           & 259.07          & 79.33          & 58.76          \\
Ours   & 50 & 50\%       & \textbf{2.53}  & 5.46  & \textbf{265.26} & \textbf{80.48} & \textbf{58.88} \\
\midrule
DDIM   & 20 & /       & 3.32           & 6.05           & 247.94          & 78.51          & 57.78          \\
Ours   & 40 & 50\%       & \textbf{2.90}  & \textbf{6.01}  & \textbf{257.47} & \textbf{79.67} & \textbf{57.97} \\
\midrule
DDIM   & 14 & /       & 5.66           & 8.80           & 218.50          & 74.57          & 55.18          \\
Ours   & 20 & 30\%       & \textbf{4.97}  & \textbf{7.30}  & \textbf{220.99} & \textbf{75.04} & \textbf{57.60} \\
\midrule
DDIM   & 10 & /       & 12.70          & 15.93          & 166.66          & 65.27          & 52.67          \\
Ours   & 20 & 50\%       & \textbf{7.00}  & \textbf{11.42} & \textbf{206.57} & \textbf{72.61} & \textbf{55.14} \\
\midrule
DDIM   & 7  & /       & 36.57          & 39.76          & 84.54           & 44.69          & 47.44          \\
Ours   & 10 & 30\%       & \textbf{16.83} & \textbf{22.76} & \textbf{143.14} & \textbf{61.05} & \textbf{50.23} \\
\bottomrule
\end{tabular}
}
\caption{
Large-DiT model results on ImageNet (cfg=1.5).
Full results included in Table 4 at Appendix A.1.
}
\label{tab:large_dit_results}
\end{table}

\begin{figure}[t] 
  \centering
  \includegraphics[width=1.0\linewidth]{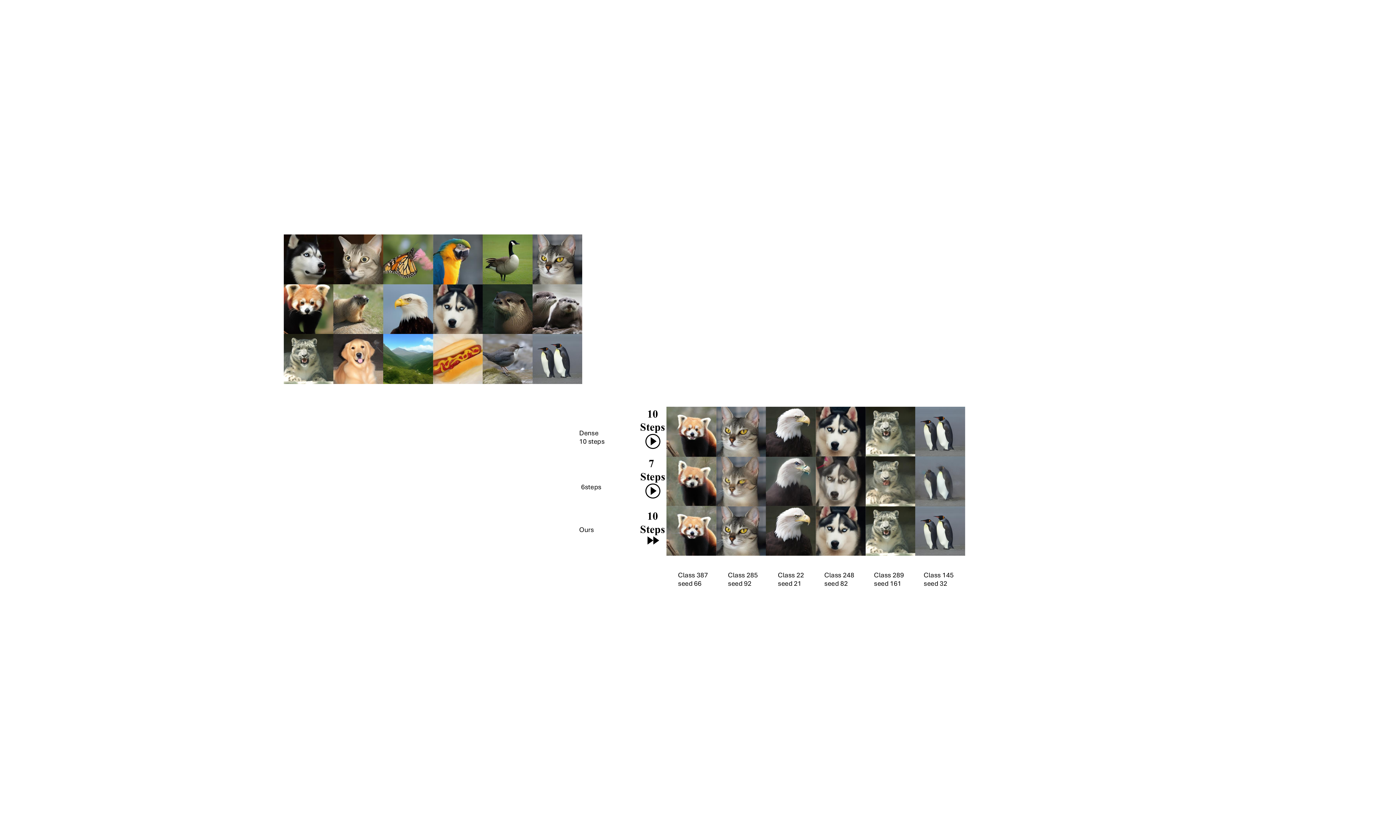}
  \caption{
  Image visualization generated by DiT-XL/2 model in 256$\times$256 resolution on mobile.
  Images at the first and second rows are generated with 10 and 7 sampling steps.
  Images at the last row are generated with 30\% lazy ratio.
  }
  \label{fig:visual_mobile}
\end{figure}

\begin{table}[!t] 
\centering
\resizebox{0.85\linewidth}{!}{
\begin{tabular}{c|c|c|c|c|c}
\toprule
\multicolumn{1}{c|}{\multirow{2}{*}{Method}} & \multicolumn{1}{c|}{\# of} & Lazy & \multicolumn{1}{c|}{\multirow{2}{*}{TMACs}}  & \multicolumn{1}{c|}{\multirow{2}{*}{IS $\uparrow$}} & Latency \\
\multicolumn{1}{c|}{} & Step & Ratio & &  & (s)  \\
\midrule
\multicolumn{6}{c}{DiT-XL/2 (256$\times$256)}             \\
\midrule
DDIM   & 50    & /      & 5.72  & 241.01 & 21.62        \\
\midrule
DDIM   & 40    & /      & 4.57  & 236.26 & 17.47        \\
DDIM   & 25    & /      & 2.86  & 230.95 & 11.33        \\
Ours   & 50    & 50\%   & 2.87  & \textbf{237.03} &      11.41   \\ 
\midrule
DDIM   & 20    & /      & 2.29  & 222.87 &     9.29    \\
DDIM   & 16    & /      & 1.83  & 211.30 &     7.60    \\
Ours   & 20    & 20\%   & 1.83  & \textbf{227.63} &       7.67  \\ 
\midrule
DDIM   & 8     & /      & 0.92  & 118.69 &     3.87    \\
DDIM   & 7     & /      & 0.80  & 91.67 &     3.54    \\
Ours   & 10    & 30\%   & 0.80  & \textbf{136.81} &      3.57   \\ 
\midrule
\multicolumn{6}{c}{DiT-XL/2 (512$\times$512)}             \\
\midrule
DDIM   & 50    & /      & 22.85 & 205.01 &     75.09    \\
\midrule
DDIM   & 40    & /      & 18.29 & 200.24 &     62.64    \\
DDIM   & 25    & /      & 11.43 & 192.71 &     41.37    \\
Ours   & 50    & 50\%   & 11.48 & \textbf{200.93} &       41.51  \\ 
\midrule
DDIM   & 13    & /      & 5.94  & 156.82 &    23.71     \\
DDIM   & 10    & /      & 4.57  & 129.19 &    19.56     \\
Ours   & 20    & 50\%   & 4.59  & \textbf{160.30} &      19.77   \\ 
\midrule
DDIM   & 9    & /      & 4.10  & 114.85 &    16.98     \\
DDIM   & 8     & /      & 3.66  & 100.75 &    15.69     \\
Ours   & 10    & 20\%   & 3.67  & \textbf{123.65} &       15.79 \\ 
\bottomrule
\end{tabular}
}
\caption{
Latency results on mobile devices with similar task performance or computation cost compared to the DDIM.
}
\label{tab:latency_results}
\end{table}

\begin{figure*}[!ht]
  \centering
  \includegraphics[width=1.0\linewidth]{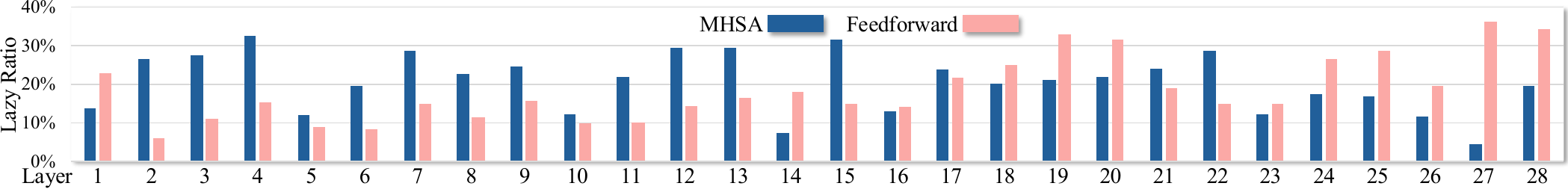}
  \caption{
  Visualization for the laziness in MHSA and Feedforward at each layer generated through DDIM 20 steps on DiT-XL.
  }
  \label{fig:visualization_sparsity_layerwise}
\end{figure*}

\subsection{Results on ImageNet}

We present the results generated with DiT officially released models compared to DDIM in Table~\ref{tab:main_results}.
Full results with more model sizes and lazy ratios are included in Table 5 of Appendix A.1. 
Due to the addition of lazy learning layers, the computational cost of our method is slightly higher than that of DDIM.
Our experiments demonstrate that our method can perform better than the DDIM on DiT models with 256$\times$256 and 512$\times$512 resolutions.
Particularly, for sampling steps fewer than 10, our method demonstrates a clear advantage over DDIM at both resolutions, highlighting the promise of our approach.
For larger models with 3B and 7B parameters, we present the results in Table~\ref{tab:large_dit_results}.
Compared to the DiT-XL/2 model with 676M parameters, Large-DiT models with a few billion parameters exhibit more redundancy during the diffusion process.
Full results for Large-DiT models are included in Table 4 at Appendix A.1.
Experiments demonstrate that at 50\% lazy ratio, our method significantly outperforms the approach of directly reducing sampling steps with DDIM.
We further visualize the images generation results
in Figure~\ref{fig:visualization_main}.
We also compare with other cache-based method Learn2Cache~\cite{ma2024learningtocache} which adopts input independent cache strategy and requires full training on ImageNet, the results are in Table 7 at Appendix A.4.
For each sampling step, Learn2Cache only has one cache strategy, whereas our method outperforms it with less training cost, demonstrating both the effectiveness and the flexibility of our method.

\subsection{Generation on Mobile}

We present the latency profiling results on mobile devices in Table~\ref{tab:latency_results}.
Our method achieves better performance with less computation cost compared to DDIM.
Additionally, when computational costs and latency are similar, our method significantly outperforms DDIM in terms of performance. 
Notably, with 10 sampling steps and a 30\% lazy ratio, our method produces significantly higher image quality in 256$\times$256 resolution than the DDIM sampler. 
Besides, we visualize the images generated on mobile in Figure~\ref{fig:visual_mobile}.
The images in the last row, generated with our method, exhibit higher quality compared to the second row, which are generated without the laziness technique under similar latency.
Therefore, our method is especially beneficial for deploying diffusion transformers on mobile devices, offering a promising solution for real-time generation on edge platforms in the future.
Meanwhile, the latency results tested on GPUs are included in Table 6 at Appendix A.2.
Our method delivers much better performance with faster latency on GPUs, especially when the number of sampling steps are fewer than 10.
Moreover, with almost the same latency, our method performs much better than DDIM.

\subsection{Ablation Study}\label{sec:ablation_study}

\paragraph{Individual Laziness.}

\begin{figure}[t] 
  \centering
  \includegraphics[width=1.0\linewidth]{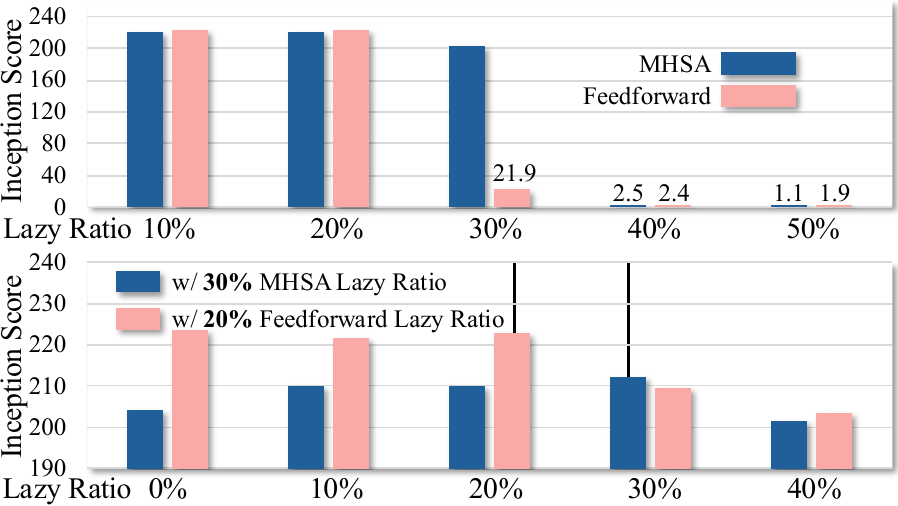}
  \caption{
  Upper figure: ablation for the generation performance with different individual laziness applied to each module independently.
  Lower figure: ablation for the generation performance with variant lazy ratio for one module and fixed lazy ratio for another module.
  }
  \label{fig:ablation_two_figure}
\end{figure}

We perform ablation studies on the laziness of MHSA and Feedforward modules separately by regulating the corresponding penalty ratios to determine the maximum applicable laziness for each, thereby exploring the redundancy within both components.
We present the results generated with DDIM 20 steps on DiT-XL/2 (256$\times$256) in the upper figure in Figure~\ref{fig:ablation_two_figure}.
The analysis indicates that the maximum applicable lazy ratio is 30\% for MHSA and 20\% for Feedforward modules.
The identification reveals that applying laziness individually to either MHSA or Feedforward network is not the most effective lazy strategy, which motivates us to apply the laziness to both modules simultaneously in our experiments.


\paragraph{Lazy Strategy.}

To optimize laziness in both MHSA and Feedforward modules for optimal performance, we fix the laziness in one module and regulate the penalty ratio of the other, varying the lazy ratio from 0\% to 40\%.
Specifically, we separately fix 30\% lazy ratio to MHSA or 20\% lazy ratio to Feedforward modules, and analyzed the model performance by regulating the lazy ratio of another module with DDIM 20 steps on DiT-XL/2 (256$\times$256).
The results, as presented in the lower figure of Figure~\ref{fig:ablation_two_figure}, reveal that the model achieves optimal performance when the same lazy ratio is applied to both MHSA and Feedforward.
Thus, we adopt the same penalty ratio for both modules in our experiments to achieve the best performance.

\paragraph{Layer-wise Laziness.}

To investigate the layer-wise importance during the diffusion process, we examined the laziness of each layer over 20 sampling steps with DiT-XL/2 model in 256$\times$256 resolution with 8 images.
The results, visualized in Figure~\ref{fig:visualization_sparsity_layerwise}, illustrate the layer-wise lazy ratio distribution and highlight key patterns in layer importance.
The analysis reveals that, for MHSA, the latter layers are more critical, whereas for Feedforward layers, the initial layers hold greater importance.
This is evidenced by the decreasing lazy ratio in MHSA and the increasing lazy ratio in MLP as going deeper.
Moreover, all layers contribute to the process, as there is no such layer that has a 100\% lazy ratio, meaning no layer is completely bypassed.
Therefore, strategies such as removing layers or optimizing model structure are not applicable for transformer-based diffusion models.


\section{Conclusion and Limitation}

In this work, we introduce the LazyDiT framework to accelerate transformer-based diffusion models.
We first show lower bound of similarity between consecutive steps is notably high.
We then incorporate a lazy skip strategy inspired by the Taylor expansion of similarity.
Experimental results validate the effectiveness of our method and we further implement our method on mobile devices, achieving better performance than DDIM.
For the limitation, there is additional computation overhead for lazy learning layers.

\bibliography{aaai25}



\clearpage

\clearpage

\appendix


\begin{center}
    \noindent\textbf{\huge Appendix}
\end{center}

\section{More Results}

\subsection{Full Results of Diffusion Transformers}\label{app:sec:full_result}

We present the full results with Large-DiT models in Table~\ref{tab:dit_results_full_appendix} and DiT models including DiT-XL and DiT-L in Table~\ref{tab:large_dit_full_results_appendix}.
A diverse range of models with varying sampling steps has confirmed the effectiveness of our method.
Besides, as the number of sampling steps decreases, our method proves to be even more effective compared to the standard 50-step sampling with the DDIM sampler.
As there is no officially released DiT-L model, thus we pretrain it for one million steps.
The results with DiT-L model here to show the generalization of our method to smaller models.

\begin{table}[h]
\centering
\resizebox{1.0\linewidth}{!}{
\begin{tabular}{c|c|c|ccccc}
\toprule
\multicolumn{1}{c|}{\multirow{2}{*}{Method}} & \multicolumn{1}{c|}{\# of} & Lazy  & \multicolumn{1}{c}{FID} & \multicolumn{1}{c}{sFID} & \multicolumn{1}{c}{IS} & \multicolumn{1}{c}{Prec.} & \multicolumn{1}{c}{Rec.} \\
\multicolumn{1}{c|}{} & Step & Ratio & $\downarrow$ & $\downarrow$ & $\uparrow$ & $\uparrow$ & $\uparrow$ \\
\midrule
\multicolumn{7}{c}{Large-DiT-3B (256$\times$256)}                                                                          \\
\midrule
DDIM   & 50 & /       & 2.10           & 4.36           & 263.83          & 80.36          & 59.55          \\
\midrule
DDIM   & 35 & /       & 2.23           & 4.48           & 262.22          & 80.08          & 60.33          \\
Ours   & 50 & 30\%       & \textbf{2.12}  & \textbf{3.32}  & \textbf{262.27} & \textbf{80.27} & 60.01          \\
\midrule
DDIM   & 25 & /       & 2.75           & 4.95           & 247.68          & 79.12          & 58.88          \\
Ours   & 50 & 50\%       & \textbf{2.42}  & \textbf{4.86}  & \textbf{257.59} & \textbf{79.71} & \textbf{59.41} \\
\midrule
DDIM   & 20 & /       & 3.46           & 5.57           & 239.18          & 77.87          & 58.71          \\
Ours   & 40 & 50\%       & \textbf{2.79}  & \textbf{5.15}  & \textbf{250.84} & \textbf{78.84} & \textbf{59.42} \\
\midrule
DDIM   & 14 & /       & 5.84           & 7.80           & 211.13          & 73.96          & 56.32          \\
Ours   & 20 & 30\%       & \textbf{4.64}  & \textbf{6.35}  & \textbf{220.48} & \textbf{75.61} & \textbf{57.81} \\
\midrule
DDIM   & 10 & /       & 13.05          & 14.17          & 162.22          & 64.89          & 51.92          \\
Ours   & 20 & 50\%       & \textbf{7.36}  & \textbf{10.55} & \textbf{197.67} & \textbf{72.14} & \textbf{54.40} \\
\midrule
DDIM   & 9  & /       & 17.92          & 18.06          & 141.22          & 59.82          & 51.28          \\
Ours   & 10 & 10\%       & \textbf{15.30} & \textbf{12.55} & \textbf{150.35} & \textbf{62.68} & \textbf{53.86} \\
\midrule
DDIM   & 7  & /       & 37.33          & 35.02          & 84.68           & 44.52          & 47.17          \\
Ours   & 10 & 30\%       & \textbf{16.40} & \textbf{12.72} & \textbf{143.70} & \textbf{61.31} & \textbf{54.17} \\
\midrule
DDIM   & 5  & /       & 78.62          & 77.09          & 28.77           & 22.19          & 43.75          \\
Ours   & 10 & 50\%       & \textbf{45.99} & \textbf{48.67} & \textbf{67.63}  & \textbf{38.97} & \textbf{47.13} \\
\midrule
\multicolumn{7}{c}{Large-DiT-7B (256$\times$256)}                                                                          \\
\midrule
DDIM   & 50 & /       & 2.16           & 4.64           & 274.89          & 80.87          & 60.00          \\
\midrule
DDIM   & 35 & /       & 2.29           & 4.83           & 267.31          & 80.42          & 59.21          \\
Ours   & 50 & 30\%       & \textbf{2.13}  & \textbf{4.49}  & \textbf{267.37} & \textbf{80.55} & \textbf{60.76} \\
\midrule
DDIM   & 25 & /       & 2.76           & 5.36           & 259.07          & 79.33          & 58.76          \\
Ours   & 50 & 50\%       & \textbf{2.53}  & 5.46  & \textbf{265.26} & \textbf{80.48} & \textbf{58.88} \\
\midrule
DDIM   & 20 & /       & 3.32           & 6.05           & 247.94          & 78.51          & 57.78          \\
Ours   & 40 & 50\%       & \textbf{2.90}  & \textbf{6.01}  & \textbf{257.47} & \textbf{79.67} & \textbf{57.97} \\
\midrule
DDIM   & 16 & /       & 4.45           & 7.28           & 232.91          & 76.30          & 56.42          \\
Ours   & 20 & 20\%       & \textbf{3.78}  & \textbf{5.45}  & \textbf{234.04} & \textbf{76.95} & \textbf{58.58} \\
\midrule
DDIM   & 14 & /       & 5.66           & 8.80           & 218.50          & 74.57          & 55.18          \\
Ours   & 20 & 30\%       & \textbf{4.97}  & \textbf{7.30}  & \textbf{220.99} & \textbf{75.04} & \textbf{57.60} \\
\midrule
DDIM   & 10 & /       & 12.70          & 15.93          & 166.66          & 65.27          & 52.67          \\
Ours   & 20 & 50\%       & \textbf{7.00}  & \textbf{11.42} & \textbf{206.57} & \textbf{72.61} & \textbf{55.14} \\
\midrule
DDIM   & 9 & /        & 17.42          & 23.24          & 143.76          & 60.14          & 51.09          \\
Ours   & 10 & 10\%       & \textbf{16.47} & \textbf{20.50} & \textbf{145.04} & \textbf{61.33} & 49.84          \\
\midrule
DDIM   & 7  & /       & 36.57          & 39.76          & 84.54           & 44.69          & 47.44          \\
Ours   & 10 & 30\%       & \textbf{16.83} & \textbf{22.76} & \textbf{143.14} & \textbf{61.05} & \textbf{50.23} \\
\bottomrule
\end{tabular}
}
\caption{
LargeDiT model results on ImageNet (cfg=1.5).
}
\label{tab:large_dit_full_results_appendix}
\end{table}

\begin{table}[!ht]
\centering
\resizebox{1.0\linewidth}{!}{
\begin{tabular}{c|c|c|ccccc}
\toprule
\multicolumn{1}{c|}{\multirow{2}{*}{Method}} & \multicolumn{1}{c|}{\# of} & Lazy  & \multicolumn{1}{c}{FID} & \multicolumn{1}{c}{sFID} & \multicolumn{1}{c}{IS} & \multicolumn{1}{c}{Prec.} & \multicolumn{1}{c}{Rec.} \\
\multicolumn{1}{c|}{} & Step & Ratio & $\downarrow$ & $\downarrow$ & $\uparrow$ & $\uparrow$ & $\uparrow$ \\
\midrule
\multicolumn{8}{c}{DiT-XL/2 (256$\times$256)}                                                        \\
\midrule
DDIM   & 50        & /  & 2.34           & 4.33           & 241.01          & 80.13          & 59.55          \\
\midrule
DDIM   & 40        & /  & 2.39           & 4.28           & 236.26          & 80.10          & 59.41          \\
Ours   & 50        & 20\%  & \textbf{2.37}  & 4.33  & \textbf{239.99} & \textbf{80.19} & \textbf{59.63} \\
\midrule
DDIM   & 35        & /  & 2.47           & 4.29           & 235.18          & 79.63          & 59.20          \\
Ours   & 50        & 30\%  & \textbf{2.46}  & \textbf{4.27}  & \textbf{237.43} & \textbf{79.99}          & \textbf{59.29}          \\
\midrule
DDIM   & 30        & /  & 2.66           & 4.40           & 234.74          & 79.85          & 58.96          \\
Ours   & 50        & 40\%  & \textbf{2.63}  & \textbf{4.35}  & \textbf{235.69} & 79.59          & 58.94          \\
\midrule
DDIM   & 25        & /  & 2.95           & 4.50           & 230.95          & 79.49          & 58.44          \\
Ours   & 50        & 50\%  & \textbf{2.70}  & \textbf{4.47}  & \textbf{237.03} & \textbf{79.77} & \textbf{58.65} \\
\midrule
DDIM   & 20        & /  & 3.53           & 4.91           & 222.87          & 78.43          & 57.12          \\
Ours   & 40        & 50\%  & \textbf{2.95}  & \textbf{4.78}  & \textbf{234.10} & \textbf{79.61} & \textbf{57.99} \\
\midrule
DDIM   & 18        & /  & 3.98	& 5.22	& 216.90 &	77.60 &	56.81 \\
Ours   & 20        & 10\%  & \textbf{3.39}  & \textbf{4.81}  & \textbf{229.32} & \textbf{79.21} & \textbf{57.66} \\
\midrule
DDIM   & 16        & /  & 4.61           & 5.69           & 211.30          & 76.57          & 56.01          \\
Ours   & 20        & 20\%  & \textbf{3.45}  & \textbf{4.83}  & \textbf{227.63} & \textbf{79.16} & \textbf{58.00} \\
\midrule
DDIM   & 14        & /  & 5.74           & 6.65           & 200.40          & 74.81          & 55.51          \\
Ours   & 20        & 30\%  & \textbf{4.44}  & \textbf{5.57}  & \textbf{212.13} & \textbf{77.11} & \textbf{56.76} \\
\midrule
DDIM   & 10        & /  & 12.05          & 11.26          & 160.73          & 66.90          & 51.52          \\
Ours   & 20        & 50\%  & \textbf{6.75}  & \textbf{8.53}  & \textbf{192.39} & \textbf{74.35} & \textbf{52.43} \\
\midrule
DDIM   & 9         & /  & 16.52 &	14.34 &	141.14 &	62.39 &	50.53 \\
Ours   & 10        & 10\%  & \textbf{12.66} & \textbf{10.70} & \textbf{158.74} & \textbf{66.05} & \textbf{51.53} \\
\midrule
DDIM   & 8         & /  & 23.17          & 19.04          & 118.69          & 56.25          & 48.52          \\
Ours   & 10        & 20\%  & \textbf{12.79} & \textbf{11.30} & \textbf{156.30} & \textbf{65.96} & \textbf{50.50} \\
\midrule
DDIM   & 7         & /  & 34.14 &	27.51 &	91.67 &	47.59 &	46.83 \\
Ours   & 10        & 30\%  & \textbf{17.05} & \textbf{13.37} & \textbf{136.81} & \textbf{62.07} & \textbf{50.37} \\
\midrule
DDIM   & 6         & /  & 52.02 &	41.77 &	60.63 &	36.21 &	44.95 \\
Ours   & 10        & 40\%  & \textbf{24.64} & \textbf{17.41} & \textbf{114.29} & \textbf{55.43} & \textbf{49.68} \\
\midrule
DDIM   & 5         & /  & 63.87          & 77.61          & 32.24           & 23.70          & 43.31          \\
Ours   & 10        & 50\%  & \textbf{37.69} & \textbf{32.34} & \textbf{84.59}  & \textbf{45.42} & \textbf{45.71} \\
\midrule
\multicolumn{8}{c}{DiT-XL/2 (512$\times$512)}                                                                                                \\
\midrule
DDIM   & 50        & / & 3.33           & 5.31           & 205.01          & 80.59          & 55.89          \\
\midrule
DDIM   & 40        & / & 3.56           & 5.45           & 200.24          & 80.45          & 54.67          \\
Ours   & 50        & 20\% & \textbf{3.54}  & \textbf{5.41}  & \textbf{201.87} & 80.41 & \textbf{55.27} \\
\midrule
DDIM   & 30        & / & 3.95           & 5.71           & 195.84          & 80.19          & 54.52          \\
Ours   & 50        & 40\% & \textbf{3.67}  & \textbf{5.65}  & \textbf{202.25} & 79.80 & \textbf{55.17} \\
\midrule
DDIM   & 25        & / & 4.26           & 6.00           & 192.71          & 79.37          & 53.99          \\
Ours   & 50        & 50\% & \textbf{3.94}  & \textbf{5.92}  & \textbf{200.93} & \textbf{80.47} & \textbf{54.05} \\
\midrule
DDIM   & 20        & /  & 5.12           & 6.60           & 184.23          & 78.37          & 53.50          \\
Ours   & 40        & 50\%  & \textbf{4.32}  & \textbf{6.50} & \textbf{196.01} & \textbf{79.64} & 53.19  \\
\midrule
DDIM   & 14        & /  & 7.87          & 8.34          & 162.48          & 74.14          & 52.21          \\
Ours   & 20        & 30\%  & \textbf{7.49}  & \textbf{8.28} & \textbf{170.31} & \textbf{75.64} & 52.05  \\
\midrule
DDIM   & 10        & /  & 14.76          & 12.38          & 129.19          & 65.51          & 48.95          \\
Ours   & 20        & 50\%  & \textbf{9.18}  & \textbf{10.85} & \textbf{160.30} & \textbf{73.16} & \textbf{49.27}  \\
\midrule
DDIM   & 8         & /  & 24.22          & 18.89          & 100.75          & 55.49          & 46.93          \\
Ours   & 10        & 20\%  & \textbf{16.28}  & \textbf{11.44} & \textbf{123.65} & \textbf{64.36} & \textbf{49.99}  \\
\midrule
\multicolumn{8}{c}{DiT-L/2 (256$\times$256)}          \\
\midrule
DDIM   & 50        & /  & 7.70           & 5.57           & 137.95          & 71.89          & 58.03          \\
\midrule
DDIM   & 40        & /  & 8.14           & 5.83           & 135.74          & 71.16          & 57.81          \\
DDIM   & 50        & 20\%  & \textbf{7.90}           & \textbf{5.51}           & \textbf{138.69}          & \textbf{71.63}          & 57.17          \\
\midrule
DDIM   & 25        & /  & 9.28           & 5.99           & 131.33          & 70.54          & 56.86          \\
DDIM   & 50        & 50\%  & \textbf{8.36}           & \textbf{5.93}           & \textbf{136.70}          & \textbf{71.15}          & 56.32          \\
\midrule
DDIM   & 20        & /  & 10.24           & 6.58           & 128.11          & 69.44          & 56.17          \\
DDIM   & 40        & 50\%  & \textbf{9.36}           & 6.77           & \textbf{132.20}          & \textbf{70.29}          & 55.23          \\
\midrule
DDIM   & 10        & /  & 25.03           & 16.66           & 89.28          & 55.83          & 50.18          \\
DDIM   & 20        & 50\%  & \textbf{18.55}           & \textbf{16.04}           & \textbf{102.84}          & \textbf{61.12}          & \textbf{51.14}          \\
\midrule
DDIM   & 8        & /  & 39.57           & 28.08           & 66.18          & 45.07          & 47.20          \\
DDIM   & 10        & 20\%  & \textbf{27.82}           & \textbf{16.71}           & \textbf{83.67}          & \textbf{53.03}          & \textbf{51.24}          \\
\bottomrule
\end{tabular}
}
\caption{
DiT model results on ImageNet (cfg=1.5).
}
\label{tab:dit_results_full_appendix}
\end{table}

\clearpage

\begin{figure*}[h]
  \centering
  \includegraphics[width=1.0\linewidth]{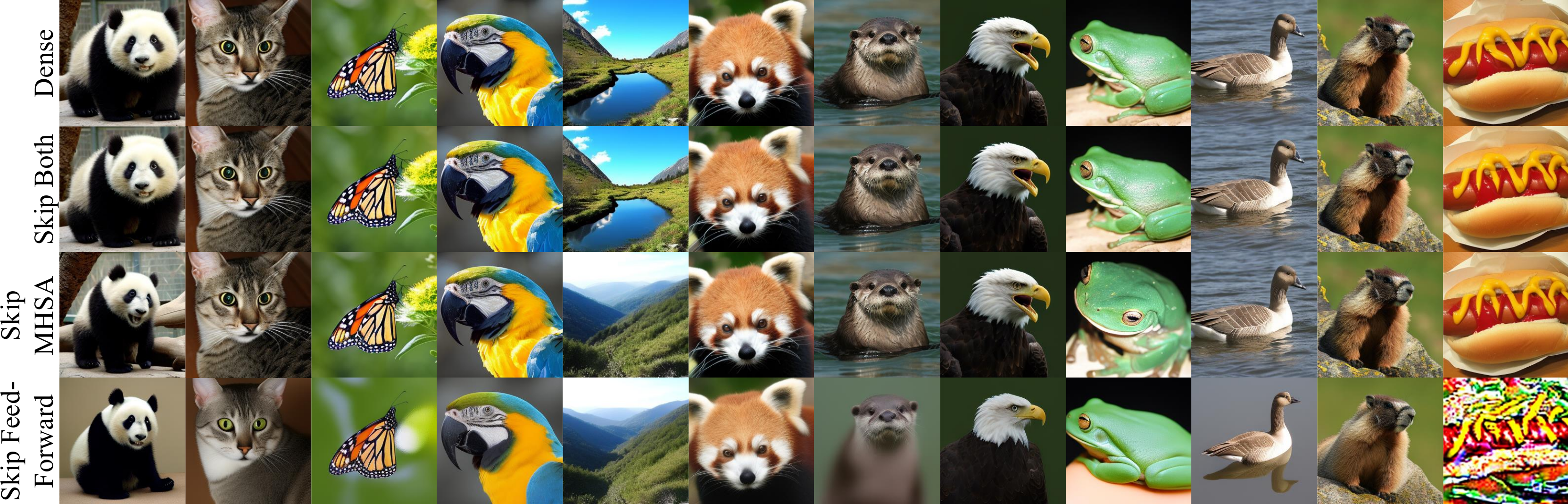}
  \caption{
  Visualization for the strategies of skipping MHSA or Feedforward only.
  }
  \label{fig:visualization_ablation_attn_or_mlp_only_skip_appendix}
\end{figure*}

\begin{figure*}[h]
  \centering
  \includegraphics[width=1.0\linewidth]{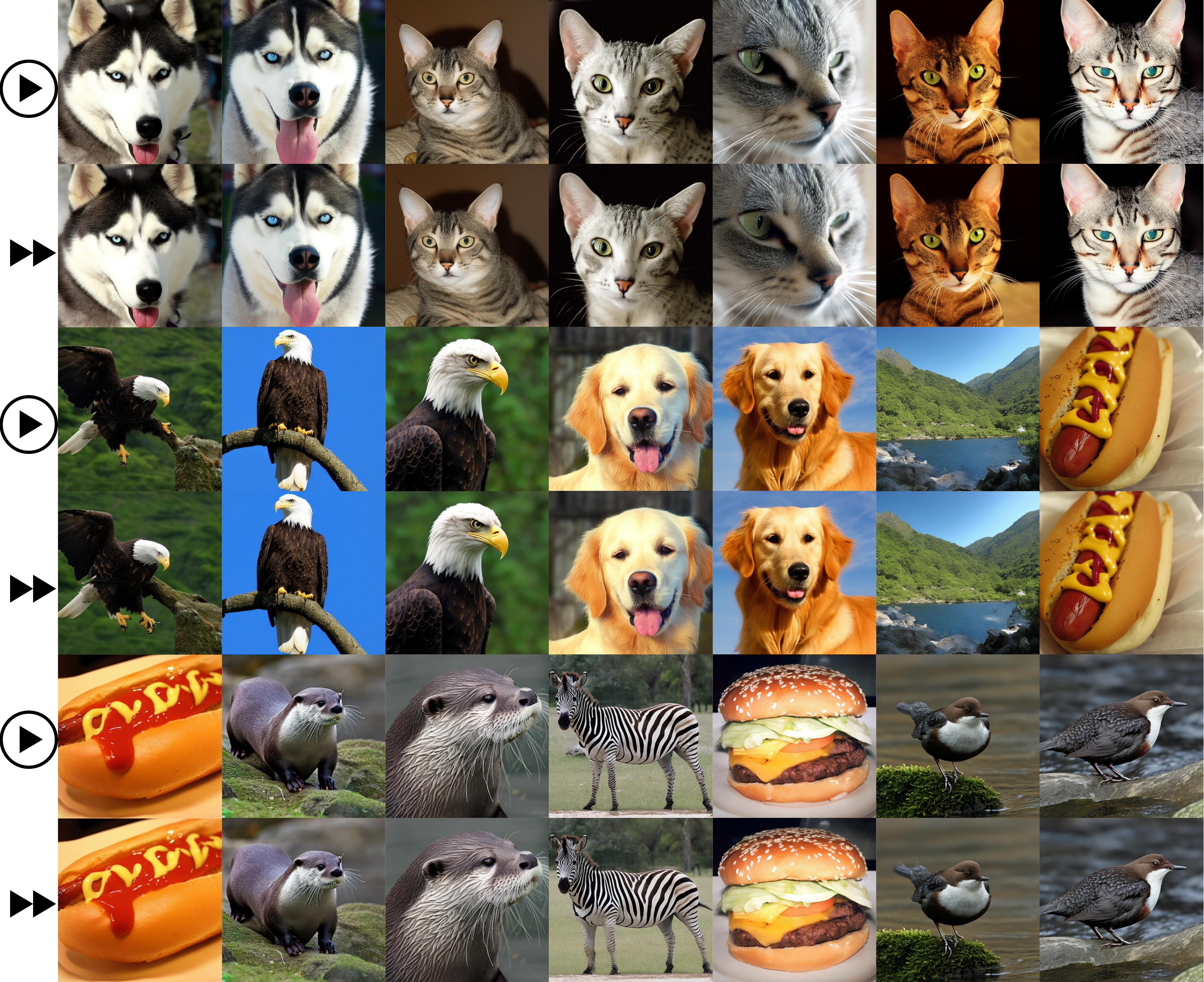}
  \caption{
  Visualization with 512$\times$512 resolution with DDIM and our method with 50\% lazy ratio.
  }
  \label{fig:visualization_512_appendix}
\end{figure*}

\subsection{Latency on GPU}\label{app:sec:gpu_latency}

We present the latency profiling results with 8 images per batch in classifier-free guidance on a single A5000 GPU in Table~\ref{tab:latency_gpu_results_appendix}.
Our method achieves better performance with faster latency on GPUs.
Meanwhile, for similar latency cost, our method outperform DDIM on the image generation quality.

\begin{table}[!t] 
\centering
\resizebox{0.85\linewidth}{!}{
\begin{tabular}{c|c|c|c|c|c}
\toprule
\multicolumn{1}{c|}{\multirow{2}{*}{Method}} & \multicolumn{1}{c|}{\# of} & Lazy & \multicolumn{1}{c|}{\multirow{2}{*}{TMACs}}  & \multicolumn{1}{c|}{\multirow{2}{*}{IS $\uparrow$}} & Latency \\
\multicolumn{1}{c|}{} & Step & Ratio & &  & (s)  \\
\midrule
\multicolumn{6}{c}{DiT-XL/2 (256$\times$256)}             \\
\midrule
DDIM   & 50    & /      & 5.72  & 241.01 & 7.39        \\
\midrule
DDIM   & 40    & /      & 4.57  & 236.26 & 5.84        \\
DDIM   & 25    & /      & 2.86  & 230.95 & 3.65        \\
Ours   & 50    & 50\%   & 2.87  & \textbf{237.03} &      3.67   \\ 
\midrule
DDIM   & 20    & /      & 2.29  & 222.87 &     2.88    \\
DDIM   & 16    & /      & 1.83  & 211.30 &     2.33    \\
Ours   & 20    & 20\%   & 1.83  & \textbf{227.63} &       2.33  \\ 
\midrule
DDIM   & 8     & /      & 0.92  & 118.69 &     1.08    \\
DDIM   & 7     & /      & 0.80  & 91.67 &     0.98    \\
Ours   & 10    & 30\%   & 0.80  & \textbf{136.81} &      1.01   \\ 
\midrule
\multicolumn{6}{c}{DiT-XL/2 (512$\times$512)}             \\
\midrule
DDIM   & 50    & /      & 22.85 & 205.01 &     38.42    \\
\midrule
DDIM   & 40    & /      & 18.29 & 200.24 &     30.68    \\
DDIM   & 25    & /      & 11.43 & 192.71 &     19.15    \\
Ours   & 50    & 50\%   & 11.48 & \textbf{200.93} &       19.23  \\ 
\midrule
DDIM   & 13    & /      & 5.94  & 156.82 &    9.74     \\
DDIM   & 10    & /      & 4.57  & 129.19 &    7.56     \\
Ours   & 20    & 50\%   & 4.59  & \textbf{160.30} &      7.60   \\ 
\midrule
DDIM   & 9    & /      & 4.10  & 114.85 &     6.69     \\
DDIM   & 8     & /      & 3.66  & 100.75 &    6.08     \\
Ours   & 10    & 20\%   & 3.67  & \textbf{123.65} &       6.11 \\ 
\bottomrule
\end{tabular}
}
\caption{
Latency results on A5000 with 8 images per batch.
}
\label{tab:latency_gpu_results_appendix}
\end{table}

\subsection{Compare to Other Cache-Base Method}\label{app:sec:compare_to_learn2cache}
We further compare our method to the Learn2Cache method~\cite{ma2024learningtocache}, we reproduce the training and evaluation with the given hyperparameter recipes.
Notably, Learn2Cache requires a full epoch of training on ImageNet, which is significantly more intensive compared to our method, which only requires 500 training steps.
Meanwhile, Learn2Cache does not support bigger laziness, offering only a single caching strategy for each specified number of sampling steps.
The results are shown in Table~\ref{tab:compare_learntocache}.
Our results achieve better performance than Learn2Cache method in most cases.
Especially, our method performs much better with less computation cost when the resolution becomes 512$\times$512.

\begin{table}[t] 
\centering
\resizebox{1.0\linewidth}{!}{
\begin{tabular}{c|c|c|cccc}
\toprule
\multicolumn{1}{c|}{\multirow{2}{*}{Method}} & \multicolumn{1}{c|}{\# of} & \multicolumn{1}{c|}{\multirow{2}{*}{TMACs}}  & \multicolumn{1}{c}{FID} &  \multicolumn{1}{c}{IS} & \multicolumn{1}{c}{Prec.} & \multicolumn{1}{c}{Rec.} \\
\multicolumn{1}{c|}{} & Step &  & $\downarrow$ & $\uparrow$ & $\uparrow$ & $\uparrow$ \\
\midrule
\multicolumn{7}{c}{DiT-XL/2 (256$\times$256)}                                                        \\
\midrule
DDIM   & 50        & 5.72  & 2.34                      & 241.01          & 80.13          & 59.55          \\
\midrule
DDIM   & 40        & 4.57  & 2.39                      & 236.26          & 80.10          & 59.41          \\
Learn2Cache & 50 & 4.36 & 2.39  & 238.89 & 80.18 & 58.59 \\
Ours   & 50        & 4.58  & \textbf{2.37}    & \textbf{239.99} & \textbf{80.19} & \textbf{59.63} \\
\midrule
DDIM   & 16        & 1.83  & 4.61                     & 211.30          & 76.57          & 56.01          \\
Learn2Cache & 20 & 1.78 & 3.47  & 227.22 & 79.15 & 56.18 \\
Ours   & 20        & 1.83  & \textbf{3.45}    & \textbf{227.63} & \textbf{79.16} & \textbf{58.00} \\
\midrule
DDIM   & 9         & 1.03  & 16.52	 &	141.14 &	62.39 &	50.53 \\
Learn2Cache & 10 & 1.04 & 12.77  & 156.39 & 66.01 & 52.26 \\
Ours   & 10        & 1.03  & \textbf{12.66}  & \textbf{158.74} & \textbf{66.05} & 51.53 \\
\midrule
\multicolumn{7}{c}{DiT-XL/2 (512$\times$512)}                                                                                                \\
\midrule
DDIM   & 50        & / & 3.33                   & 205.01          & 80.59          & 55.89          \\
\midrule
DDIM   & 30        & 13.71 & 3.95                     & 195.84          & 80.19          & 54.52          \\
Learn2Cache & 50 & 14.19 & 3.76  & 201.13 & 80.08 & 54.21 \\
Ours   & 50        & 13.77 & \textbf{3.67}    & \textbf{202.25} & 79.80 & \textbf{55.17} \\
\bottomrule
\end{tabular}
}
\caption{
DiT model results on ImageNet (cfg=1.5) compared to Learn2Cache method~\cite{ma2024learningtocache}.
}
\label{tab:compare_learntocache}
\end{table}

\subsection{Ablation for Skipping MHSA or Feedforward}\label{app:sec:ablation_mhsa_feedforward}

We present the ablation study for skipping either MHSA or Feedforward layers individually, utilizing the sparse learning weights trained with the combined skipping of both components at same sparsity.
The results are shown in Figure~\ref{fig:visualization_ablation_attn_or_mlp_only_skip_appendix}.

\subsection{More Visualization}

More visualization is provided in Figure~\ref{fig:visualization_512_appendix} in 512$\times$512 resolution compared to DDIM sampler with 50\% lazy ratio.


\onecolumn

\section{Theoretical Preliminary}

In this section, we provide the preliminary of our theoretical results.

\subsection{Notations}\label{sec:preli:notations}
For two vectors $x \in \mathbb{R}^n$ and $y \in \mathbb{R}^n$, we use $\langle x, y \rangle$ to denote the inner product between $x,y$, i.e., $\langle x, y\rangle = \sum_{i=1}^n x_i y_i$.
We use $e_i$ to denote a vector where only $i$-th coordinate is $1$, and other entries are $0$.
For each $a, b \in \R^n$, we use $a \circ b \in \R^n$ to denote the Hardamard product, i.e. the $i$-th entry of $(a \circ b)$ is $a_i b_i$ for all $i \in [n]$.
We denote the $i$-th row of a matrix $X$ as $X_i$. 
We use ${\bf 1}_n$ to denote a length-$n$ vector where all the entries are ones.
We use $x_{i,j}$ to denote the $j$-th coordinate of $x_i \in \mathbb{R}^n$.
We use $\|x\|_p$ to denote the $\ell_p$ norm of a vector $x \in \mathbb{R}^n$, i.e. $\|x\|_1 := \sum_{i=1}^n |x_i|$, $\|x\|_2 := (\sum_{i=1}^n x_i^2)^{1/2}$, and $\|x\|_{\infty} := \max_{i \in [n]} |x_i|$.
For $k > n$, for any matrix $A \in \mathbb{R}^{k \times n}$, we use $\|A\|$ to denote the spectral norm of $A$, i.e. $\|A\|:=\sup_{x\in \mathbb{R}^n} \|Ax\|_2 / \|x\|_2$.
We define $\| A \|_{\infty} := \max_{i \in [m], j \in [n]} |A_{i,j}|$ for $A \in \R^{m \times n}$.
For a square matrix $A$, we use $\tr[A]$ to denote the trace of $A$, i.e., $\tr[A] = \sum_{i=1}^n A_{i,i}$.
For two matrices $X, Y$, the standard inner product between matrices is defined by $\langle X, Y \rangle := \tr[X^\top Y]$. 
We define  $\| X \|_F := (\sum_{i=1}^m \sum_{j=1}^n  X_{i,j}^2)^{1/2}$ for a matrix $X \in \mathbb{R}^{m \times n}$. 


\subsection{Facts}\label{sec:theory:fact}
In this section, we provide several facts we use.
\begin{fact}
\label{fac:matrix_norm_infty}
For a matrix $A \in \R^{m \times n}$, we have
\begin{align*}
    \|A\|_\infty \leq \|A\| \leq \sqrt{mn}\|A\|_\infty.
\end{align*}
\end{fact}

\begin{fact}
\label{fac:matrix_norm_F}
For a matrix $A \in \R^{m \times n}$, we have
\begin{align*}
    \|A\| \leq \| A \|_F \leq \sqrt{k} \| A \|,
\end{align*}
where $k$ is the rank of $A$. 
\end{fact}

\begin{fact}[Trace]
\label{fac:fact_trace}
We have the following properties of trace:
\begin{itemize}
    \item $\| X\|_F^2 = \tr[X^\top X]$ for matrix $X \in \R^{m \times n}$.
    \item $\tr[X^\top Y] = \tr[X Y^\top]$ for matrices $X, Y \in \R^{m \times n}$.
    \item $\tr[A B C] = \tr[C A B] = \tr[B C A]$ for matrices $A, B, C \in \R^{m \times m}$.
\end{itemize}
\end{fact}

\begin{fact}
\label{fac:matrices_similarity}
For matrices $A, B \in \R^{m \times n}$, we have 
\begin{align*}
    \| A - B\|_F^2 = \| A \|_F^2 + \| B \|_F^2 - 2 \tr [A^\top B].
\end{align*}
\end{fact}
\begin{proof}
    We can show 
\begin{align*}
    \| A - B \|_F^2 = & ~ \tr[(A - B)^\top (A - B)] \\
    = & ~ \tr[A^\top A - A^\top B - B^\top A + B^\top B]\\
    = & ~ \tr[A^\top A] + \tr[B^\top B] - 2\tr[A^\top B]\\
    = & ~ \| A \|_F^2 + \| B \|_F^2 - 2\tr[A^\top B]
\end{align*}
where the first step follows from Fact~\ref{fac:fact_trace}, the second step follows from the basic algebra, the third and the last step follow from Fact~\ref{fac:fact_trace}.
\end{proof}


\begin{fact}[Taylor expansion for matrix]
\label{fac:matrix_tylar_expand}
Here, we provide the Taylor expansion for function $f(\cdot): \R^{m \times n} \to \R^{m \times n} $. For matrices $X \in \R^{m \times n}$, suppose we expand $X$ around $X_0$. 
We have
\begin{align*}
    f(X) = f(X_0) + J \cdot (X- X_0)  + O(\| X - X_0 \|^2)
\end{align*}
where $J$ is a Jacobian matrix.
\end{fact}

\subsection{Definitions}
In this section, we provide key definitions we use for the proofs.
We define the attention module and feedforward module as follows.

\begin{definition}[Self-attention module]\label{def:self_attention}  
Given input matrix $X \in \R^{N \times D}$ and weight matrices $W_Q, W_K, W_V \in \R^{D \times D}$, (which represent the query, key and value weight matrices separately), we have self-attention module at $l$-th layer defined by:
\begin{align*}
    \mathcal{F}_l^\attn(X) := \underbrace{D^{-1}}_{N \times N} \underbrace{A}_{N \times N} \underbrace{V}_{N \times D}    
\end{align*}
where (1) $V := X W_V$, (2) $W := W_Q W_K^\top$, (3) $A:= \exp (X W X^\top)$, and (4) $D := \diag (A \cdot  {\bf 1}_n)$.
\end{definition}

\begin{definition}[Feedforward module]\label{def:feedforward}
Given input matrix $X \in \R^{N \times D}$, we define the feedforward module at $l$-th layer as
\begin{align*}
    \mathcal{F}_l^\feed(X) := \underbrace{X}_{N \times D} \underbrace{W^\feed}_{D \times D}
\end{align*}
where $W^\feed \in \R^{D \times D}$ is the linear weight for the feedforward layer.
\end{definition}


\section{Theoretical Results}
For simplicity of proofs, we assume the batch size $B=1$. This will only affect the results by a constant factor of $B$.
In this section, we provide the theoretical part of our paper.

\subsection{More Related Work}

\paragraph{Lazy Update.}
The concept of lazy update has emerged as a powerful strategy in addressing computational challenges. 
Lazy updating defers certain computations or updates until they are absolutely necessary, which can significantly reduce computational overhead in large-scale problems.  
This approach has applications to several fundamental theoretical frameworks such as Linear Programming (LP) and Semi-Definite Programming (SDP)~\cite{a00,dgj+06,aw08,dkk+16,dhl19,cls19,jlt20,jkl+20,gs22,syz23}.  
The application of lazy update strategies extends beyond LP and SDP to other important areas in computer science and machine learning. 
Dynamic maintenance of data structures and algorithms is one such area where lazy updates have proven valuable \cite{cls19,lsz19,b20,jlsw20,blss20,jkl+20,jswz21,sy21,dly21,b21,hjs+22,gs22}. 
This approach allows for efficient updates to solutions as input data changes, without the need for complete recomputation.
In the realm of machine learning, Empirical Risk Minimization (ERM) is a fundamental principle that has benefited from lazy update techniques. 
ERM has been extensively studied in various contexts \cite{n83,v91,pj92,bbm05,bb08,njls09,mb11,fgrw12,n13,jz13,v13,sz13,db14,fgks15,zx17,jlgj18,lsz19,lszz23,bsy23}, with recent work focusing on improving its efficiency through lazy updates.
Support Vector Machines (SVMs), a popular machine learning model, have also seen advancements through the application of lazy update strategies. Recent research has explored ways to optimize SVM training and inference using these techniques \cite{cl01,j06,tlto23,bsz23,lsz23,gms23,gsz23,gswy23}, leading to improved performance in both time and space complexity.

\paragraph{Lazy Machine Learning.}
Lazy update strategies have been extensively explored in the context of machine learning to enhance algorithmic efficiency by reducing iterative loops. 
For instance, \cite{sjr+18} proposed a memory-based parameter adaptation (MbPA) method, leveraging lazy updates to optimize model parameters. 
In the domain of reinforcement learning, \cite{jfpg22} introduced lazy-MDPs, incorporating lazy update techniques to streamline the learning process. 
Furthermore, \cite{jtt+22} investigated the interplay between forgetting and privacy protection during model training, highlighting the potential of lazy updates in balancing computational efficiency and data privacy. 
\cite{yrs+21} proposed a meta-learning approach via in-context tuning, demonstrating the effectiveness of lazy update principles in adapting models to new tasks. 
Recently, \cite{mfn+23} provided a comprehensive overview of the shortcut learning problem in large language models for natural language understanding tasks, emphasizing the role of lazy update strategies in mitigating this challenge. 

\paragraph{Attention Theory.}
Attention mechanisms have been extensively studied and developed over the years, becoming a fundamental component in various neural network architectures. 
\cite{dls23,lsx+23,gll+24a} investigate the softmax regression problem, while \cite{syz23} propose and analyze the attention kernel regression problem. 
The rescaled hyperbolic functions regression is examined by \cite{gsy23}.
Following this, \cite{swy23,lswy23} delve into two-layer attention regression problems. 
\cite{gll+24b} demonstrate that attention layers in transformers learn two-dimensional cosine functions. 
Additionally, \cite{dsxy23} explore data recovery using attention weights, and~\cite{kmz23,sxy23} investigate the replacement of the softmax unit with a polynomial unit.
Moreover, some works theoretically explore variations or combinations of the attention mechanism with other techniques, such as quantum attention~\cite{gsyz23,gsyz24_quantum_kro}, tensor attention~\cite{as24_iclr, lssz24_tat}, and differentially private attention~\cite{gls+24_dp_tree, gsy23_dp, gls+24_ntk} and other applications such as \cite{cklm19,tdp19,hl19,vb19,b22,wsd+23, wcz+23,wxz+24,bsz23,gls+24_diffusion,cls+24,cll+24_icl,cll+24_rope,ssx23_nns,qss23_gnn,dms23_spar,llsz24_nn_tw,lss+24_relu,lls+24_io,lss+24_mutlilayer,lssz24_tat,lls+24}.

\subsection{Effect of Scaling Factor}\label{sec:effect_scaling_factor}
In this section, we examine the scaling factor used in DiT architecture.
We first prove the vector case.
\begin{lemma}[Scaling and shifting for one row]\label{lem:scaling_row}
If the following conditions hold:
\begin{itemize}
    \item Let $x_1, x_2 \in \R^D$. 
    \item Let $a, b, c \in \R^D$ and $a, b \neq 0$. 
\end{itemize}
Then, we can show that there exist $a,b,c$ such that 
\begin{align*}
    \| a  \circ x_1 + b \circ x_2 + c \|_2 \leq \eta 
\end{align*} 
where $\eta \in (0,0.1)$. 
\end{lemma}

\begin{proof}
Let $a = \frac{0.5 \eta}{\|x_1\|_2} \cdot {\bf 1}_D$, $b = \frac{0.5 \eta}{\|x_2\|_2} \cdot {\bf 1}_D$, and $c=0$. 

Then, we have
\begin{align*}
    & ~ \| a  \circ x_1 + b \circ x_2 + c \|_2 \\
    = & ~ \| \frac{0.5 \eta \cdot x_1}{\| x_1 \|_2} + \frac{0.5 \eta \cdot x_2}{\| x_2 \|_2} + 0 \|_2 \\
    \leq & ~ \| \frac{0.5 \eta \cdot x_1}{\| x_1 \|_2} \|_2 + \| \frac{0.5 \eta \cdot x_2}{\| x_2 \|_2} \|_2 \\
    = & ~ \eta
\end{align*}
where the first step follows from the way we pick $a,b,c$, the second step follows from the triangle inequality, the last step follows from basic algebra.
\end{proof}

It can be proven that for any matrix, there exist scaling and shifting factors that can adjust the input such that the distance between them is bounded.
\begin{lemma}[Scaling and shifting]\label{lem:scaling_shifting}
If the following conditions hold:
\begin{itemize}
    \item Let $X_1, X_2 \in \R^{N \times D}$.
    \item Let $a, b, c \in \R^D$ and $a, b \neq 0$. 
    \item Let $A \in \R^{N \times D}$ be defined as the matrix that all rows are $a$, i.e. $A_i := a$ for $i \in [N]$, and $B, C$ be defined as the same way. 
    \item We define $ M_1 := \max_{i \in [N]} \|X_{1,i}\|_2 $ for $X_1$.
    \item We define $ M_2 := \max_{i \in [N]} \|X_{2,i}\|_2 $ for $X_2$. 
\end{itemize}
Then, we can show that there exist $a,b,c$ such that
\begin{align*}
    \|A \circ X_1 + B \circ X_2 + C \|_F \leq \eta
\end{align*} 
where $\eta \in (0,0.1)$.
\end{lemma}

\begin{proof}

Let $a = {\bf 1}_D \cdot \frac{0.5 \eta}{N M_1}$, $b = {\bf 1}_D \cdot \frac{0.5 \eta}{N M_2}$, and $c=0$.
Let $\wt{A} = A \circ X_1$ and $\wt{B} = B \circ X_2$ and $\wt{C} = C$. 

We define
\begin{align*}
    \eta_0 := \eta / (3\sqrt{N}).
\end{align*}

We can show
\begin{align}\label{eq:wt_A}
    \| \wt{A}_i \|_2 = & ~ \| a \circ X_{1,i}\|_2 \notag\\
    = & ~ \frac{0.5 \eta}{N M_1} \| X_{1,i} \|_2 \notag\\
    \leq & ~ \frac{0.5 \eta}{N} \notag \\
    \leq & ~ \eta_0
\end{align}
where the first step follows from the definition of  $\wt{A}$, the second step follows from the definition of $a$, the third step follows from the definition of $M_1$ and $X_{1,i}$, the last step follows from the definition of $\eta_0$.

We can show 
\begin{align}\label{eq:wt_B}
    \| \wt{B}_i \|_2 = & ~ \| b \circ X_{2,i}\|_2 \notag\\
    = & ~ \frac{0.5 \eta}{N M_2} \| X_{2,i} \|_2 \notag\\
    \leq & ~ \frac{0.5 \eta}{N} \notag\\
    \leq & ~ \eta_0
\end{align}
where the first step follows from the definition of  $\wt{B}$, the second step follows from the definition of $b$, the third step follows from the definition of $M_2$ and $X_{2,i}$, the last step follows from the definition of $\eta_0$.

Because $c=0$, we can show 
\begin{align*}
    \| \wt{C}_i \|_2 = 0.
\end{align*}


Then, we can show
\begin{align*}
    \| \wt{A} + \wt{B} + \wt{C} \|_F^2 
    = & ~ \sum_{i=1}^N \| \wt{A}_i + \wt{B}_i + \wt{C}_i \|_2^2 \\
    \leq & ~ \sum_{i=1}^N 3 ( \| \wt{A}_i \|_2^2 + \| \wt{B}_i \|_2^2 + \| \wt{C}_i \|_2^2 ) \\
    \leq & ~ 3 N \cdot 3\eta_0^2 \\
    \leq & ~ \eta^2
\end{align*}
where the first step follows from definition of Frobenius norm, the second step follows from Cauchy–Schwarz inequality, the third step follows from Eq.\eqref{eq:wt_A} and Eq.\eqref{eq:wt_B}, and the last step follows from definition of $\eta_0$.

Taking the square root of the both side of the above equation, this can be further simplified as:
\begin{align*}
\| \wt{A} + \wt{B} + \wt{C} \|_F \leq \eta.
\end{align*}
\end{proof}

Using the above idea, we can prove Theorem~\ref{thm:scaling_shifting:formal}.
\begin{theorem}[Effect of scaling and shifting, formal version of Theorem~\ref{thm:scaling_shifting:informal}]\label{thm:scaling_shifting:formal}
If the following conditions hold:
\begin{itemize}
    \item Let $X_{l,t-1}^{\Phi}, X_{l,t}^{\Phi} \in \R^{N \times D}$.
    \item Let $a_t, a_{t-1}, b_t, b_{t-1} \in \R^D$. 
    \item Let $A_t \in \R^{N \times D}$ be defined as the matrix that all rows are $a_t$, i.e. $(A_t)_i := a_t$ for $i \in [N]$, and $A_{t-1}, B_t, B_{t-1}$ be defined as the same way. 
\end{itemize}
Then, we can show there exist $A_{t-1}, A_t, B_{t-1}, B_t$
\begin{align*}
    \| (A_{t-1} \circ X_{l,t-1}^{\Phi} + B_{t-1}) - (A_{t} \circ X_{l,t}^{\Phi} + B_{t}) \| \leq \eta
\end{align*}
where $\eta \in (0,0.1)$.
\end{theorem}

\begin{proof}
We can show
\begin{align*}
    & ~ \| (A_{t-1} \circ X_{l,t-1}^{\Phi} + B_{t-1}) - (A_{t} \circ X_{l,t}^{\Phi} + B_{t}) \| \\
    = & ~ \| A_{t-1} \circ X_{l,t-1}^{\Phi} - A_{t} \circ X_{l,t}^{\Phi} + B_{t-1} - B_t \| \\
    \leq & ~ \| A_{t-1} \circ X_{l,t-1}^{\Phi} - A_{t} \circ X_{l,t}^{\Phi} + B_{t-1} - B_t \|_F \\
    := & ~ \|A \circ X_{l,t-1}^{\Phi} + B \circ X_{l,t}^{\Phi} + C  \|_F \\
    \leq & ~ \eta
\end{align*}
where the first step follows from basic algebra, the second step follows from Fact~\ref{fac:matrix_norm_F}, the third step follows from we define $A := A_{t-1}, B := - A_t, C:= B_{t-1} - B_t$, the last step follows from Lemma~\ref{lem:scaling_shifting}. 
\end{proof}






\subsection{Lipschitz of Model Outputs}\label{sec:lipschitz}

We first prove the Lipschitz for attention layer.
\begin{lemma}[Lemma H.5 in \cite{dsxy23}]\label{lem:attn_lip_single_value}
If the following conditions hold:
\begin{itemize}
    \item Let $W, W_V \in \R^{D \times D}, X \in \R^{N \times D} $ be defined as in Definition~\ref{def:self_attention}.
    \item $R > 1$ be some fixed constant.
    \item Assume $\|  W \| \leq R, \| W_V \|\leq R, \| X \|\leq R$.
\end{itemize}
For $i \in [N], j \in [D]$, we have
\begin{align*}
    |\mathcal{F}_l^\attn(X)_{i,j} - \mathcal{F}_l^\attn(\wt{X})_{i,j}| 
    \leq 5 R^4 \sqrt{ND} \cdot \|X - \wt{X}\|. 
\end{align*}  
\end{lemma}
Using the previous results from \cite{dsxy23}, we can prove the Lipschitz for attention layer.
\begin{lemma}[Lipschitz of attention]\label{lem:lip_attn}
If the following conditions hold:
\begin{itemize}
    \item Let $W, W_V \in \R^{D \times D}, X \in \R^{N \times D} $ be defined as in Definition~\ref{def:self_attention}.
    \item $R > 1$ be some fixed constant.
    \item Assume $\|  W \| \leq R, \| W_V \|\leq R, \| X \|\leq R$.
\end{itemize}
Then, we have the attention module (Definition~\ref{def:self_attention}) satisfying:  
\begin{align*}
    \|\mathcal{F}_l^\attn(X) - \mathcal{F}_l^\attn(\wt{X})\|
    \leq 5 R^4 ND \cdot \|X - \wt{X}\|.
\end{align*}
\end{lemma}

\begin{proof}
Then we can show
\begin{align*}
    & ~ \|\mathcal{F}_l^\attn(X) - \mathcal{F}_l^\attn(\wt{X})\|\\
    \leq & ~ \sqrt{ND} \|\mathcal{F}_l^\attn(X) - \mathcal{F}_l^\attn(\wt{X})\|_\infty \\
    = & ~ \sqrt{ND} \max_{i \in [N], j \in [D]} |(\mathcal{F}_l^\attn(X) - \mathcal{F}_l^\attn(\wt{X}))_{i,j}| \\
    \leq & ~ 5 R^4 N D \|X - \wt{X}\|
\end{align*}
where the first step follows from Fact~\ref{fac:matrix_norm_infty}, the second step follows the definition of $\|\cdot\|_\infty$, and the second step follows from Lemma~\ref{lem:attn_lip_single_value}.
\end{proof}

We then prove the Lipschitz for feedforward layer.
\begin{lemma}[Lipschitz of feedforward]\label{lem:lip_feed}
If the following conditions hold:
\begin{itemize}
    \item Let $X \in \R^{N \times D}$ and $W^\feed \in \R^{D \times D}$ be defined in Definition~\ref{def:feedforward}.
    \item $R > 1$ be some fixed constant.
    \item Assume $\|X\| \leq R, \|W^\feed\| \leq R$.
\end{itemize}

Then, we have the feedforward module (Definition~\ref{def:feedforward}) satisfying: 
\begin{align*}
    \|\mathcal{F}_l^\feed(X) - \mathcal{F}_l^\feed(\wt{X})\| \leq R \cdot \|\ X - \wt{X} \|.
\end{align*}
\end{lemma}

\begin{proof}
We can show
\begin{align*}
    \|\mathcal{F}_l^\feed(X) - \mathcal{F}_l^\feed(\wt{X})\| = & ~\|X W^\feed - \wt{X} W^\feed \| \\
    \leq & ~ \| W^\feed\| \cdot \|\ X - \wt{X} \| \\
    \leq & ~ R \|\ X - \wt{X} \|
\end{align*}
where the first step follows from Definition~\ref{def:feedforward}, the second step follows from $\|AX\| \leq \|A\| \cdot \|X\|$ for two matrices $A,X$, and the last step follows from $\|W^\feed\| \leq R$.
\end{proof}

Combining the Lipschitz of two modules, we have the following lemma.
\begin{lemma}\label{lem:lip_attn_feed}
If the following conditions hold:
\begin{itemize}
    \item Let $X \in \R^{N \times D}$. 
    \item Let $W^\feed \in \R^{D \times D}$ be defined in Definition~\ref{def:feedforward}.
    \item Let $W, W_V \in \R^{D \times D}$ be defined as in Definition~\ref{def:self_attention}.
    \item $R > 1$ be some fixed constant.
    \item Assume $\|  W^\feed \| \leq R, \|  W \| \leq R, \| W_V \|\leq R, \| X \|\leq R$.
    \item $C = 5R^4 N D$ when $\Phi = \attn$, and $C = R$ when $\Phi = \feed$. 
\end{itemize}
Then, we have
\begin{align*}
        \|\mathcal{F}_l^\Phi(X) - \mathcal{F}_l^\Phi(\wt{X})\| \leq C \cdot \|\ X - \wt{X} \|.
\end{align*}
\end{lemma}
\begin{proof}
Combining Lemma~\ref{lem:lip_attn} and \ref{lem:lip_feed}, we finish the proof.
\end{proof}

\subsection{Similarity Lower Bound}\label{sec:similarity_lower_bound}
In this section, we prove that the similarity of outputs between two diffusion timestep can be bounded.


\begin{theorem}[Similarity lower bound, formal version of Theorem~\ref{thm:similar_or_dissimilar:informal}]\label{thm:similar_or_dissimilar:formal}
If the following conditions hold:
\begin{itemize}
    \item Let $\alpha \in (0,0.5)$. 
    \item Let $X_{l,t}^\Phi \in \R^{N \times D}$, $W^\feed \in \R^{D \times D}$ be defined in Definition~\ref{def:feedforward}, and $W, W_V \in \R^{D \times D}$ be defined as in Definition~\ref{def:self_attention}.
    \item Assume $\|  W^\feed \| \leq R, \|  W \| \leq R, \| W_V \|\leq R, \| X \|\leq R$ for some fixed constant $R > 1$.
    \item Assume $R_1 \leq \|X_{l,t-1}^\Phi - X_{l,t}^\Phi \| \leq R_2$ for $R_1,R_2 \in \R$. 
    \item $C = 5R^4 N D$ when $\Phi = \attn$, and $C = R$ when $\Phi = \feed$. 
    \item We define $Y_{l,t}^{\Phi} := \F_l^{\Phi}(X_{l,t}^\Phi)$ and assume $\| Y_{l,t}^{\Phi} \|_F=1$ for any $t$. 
\end{itemize}
Then, the similarity between $Y_{l,t-1}^{\Phi}$ and $Y_{l,t}^{\Phi}$ can be bounded by:
\begin{align*}
    f(Y_{l,t-1}^{\Phi}, Y_{l,t}^{\Phi}) \geq 1- \alpha.
\end{align*}
\end{theorem}

\begin{proof}
Suppose we use cosine similarity. 
We derive a new form of $f$ first.
\begin{align}\label{eq:similarity_f}
    & ~ f(Y_{l,t-1}^{\Phi}, Y_{l,t}^{\Phi}) \notag \\
    = & ~ \frac{\tr[Y_{l,t-1}^{\Phi} \cdot (Y_{l,t}^{\Phi})^\top]}{\|Y_{l,t-1}^{\Phi}\|_F \cdot \|Y_{l,t}^{\Phi}\|_F}  \notag \\
    = & ~ \frac{0.5 (\| Y_{l,t-1}^{\Phi} \|_F^2 + \| Y_{l,t}^{\Phi} \|_F^2 - \| Y_{l,t-1}^{\Phi} - Y_{l,t}^{\Phi} \|_F^2)}{\|Y_{l,t-1}^{\Phi}\|_F \cdot \|Y_{l,t}^{\Phi}\|_F}  \notag \\
    = & ~ 1- 0.5\cdot \| Y_{l,t-1}^{\Phi} - Y_{l,t}^{\Phi} \|_F^2 
\end{align}
where the first step follows from the definition of cosine similarity, the second step follows from Fact~\ref{fac:matrices_similarity}, the third step follows from $\| Y_{l,t}^{\Phi} \|_F=1$ for any $t$.

Then, we can show that
\begin{align*}
    & ~ f(Y_{l,t-1}^{\Phi}, Y_{l,t}^{\Phi}) \\
    = & ~ 1- 0.5 \cdot \| Y_{l,t-1}^{\Phi} - Y_{l,t}^{\Phi} \|_F^2\\
    \geq & ~ 1- 0.5\min\{N,D\} \cdot \| Y_{l,t-1}^{\Phi} - Y_{l,t}^{\Phi} \|^2\\
    \geq & ~ 1- 0.5 C^2\min\{N,D\} \cdot \|X_{l,t-1}^{\Phi} - X_{l,t}^{\Phi}\|^2\\
    \geq & ~ 1 - 0.5 C^2 R_2^2\min\{N,D\}\\
    := & ~ 1 - \alpha
\end{align*}
where the first step follows from Eq.~\eqref{eq:similarity_f}, the second step follows from Fact~\ref{fac:matrix_norm_F}, the third step follows from Lemma~\ref{lem:lip_attn_feed}, the fourth step follows from $\|X_{l,t-1}^{\Phi} - X_{l,t}^{\Phi} \| \leq R_2$, and the last step follows from we define $\alpha := 0.5 C^2 R_2^2\min\{N,D\}$.

\end{proof}

\subsection{Approximating Similarity Function}\label{sec:approx_similarity}
In this section, we prove that the similarity function can be approximated by a linear linear with certain error.
\begin{theorem}[Linear layer approximation, formal version of Theorem~\ref{thm:linear_layer:informal}]\label{thm:linear_layer:formal}
If the following conditions hold:
\begin{itemize}
    \item The similarity function $f$ is assumed to be sufficiently smooth such that its derivatives up to at least the second order are well-defined and continuous over the range of interest.
    \item We define $Y_{l,t}^{\Phi} := \F_l^{\Phi}(X_{l,t}^{\Phi})$ and assume $\| Y_{l,t}^{\Phi} \|_F=1$ for any $t$. 
\end{itemize}
There exist weights $W_l$ of a linear layer in the $l$-th layer of diffusion model such that $f(Y_{l,t-1}^{\Phi}, Y_{l,t}^{\Phi}) = \langle W^\Phi_l, X_{l,t}^{\Phi} \rangle$.
\end{theorem}

\begin{proof}
Suppose we use cosine similarity. We have
\begin{align}
    & ~ f(Y_{l,t-1}^{\Phi}, Y_{l,t}^{\Phi}) \notag \\
    = & ~ \frac{\tr[Y_{l,t-1}^{\Phi} \cdot (Y_{l,t}^{\Phi})^\top]}{\| Y_{l,t-1}^{\Phi} \|_F \cdot \|Y_{l,t}^{\Phi}\|_F} \notag \\
    = & ~ \tr[Y_{l,t-1}^{\Phi} \cdot (Y_{l,t}^{\Phi})^\top]  \notag \\
    = & ~ \tr[(Y_{l,t-1}^{\Phi})^\top \cdot Y_{l,t}^{\Phi}] \notag \\
    = & ~ \tr[(Y_{l,t-1}^{\Phi})^\top \cdot (0 + J \cdot (X_{l,t}^{\Phi} - 0) + O(1))] \label{eq:taylor_Jacobian matrix}\\
    = & ~ \tr[(Y_{l,t-1}^{\Phi})^\top \cdot J \cdot X_{l,t}^\Phi)] + O(1) \notag \\
    = & ~ \langle (Y_{l,t-1}^{\Phi})^\top \cdot J , X_{l,t}^\Phi \rangle + O(1) \notag \\
    := & ~ \langle W^\Phi_l, X_{l,t}^\Phi \rangle + O(1) \notag
\end{align}
where the first step follows from the definition of cosine similarity, the second step follows from $\| Y_{l,t}^{\Phi} \|_F=1$ for any $t$, the third steps from Fact~\ref{fac:fact_trace}, the fourth step follows from Fact~\ref{fac:matrix_tylar_expand} and we expand $Y_{l,t}^{\Phi}$ around $0$, the fifth step follows from the simple algebra, the sixth follows from the definition of inner product, the last step follows from we define $W^\Phi_l := (Y_{l,t-1}^{\Phi})^\top \cdot J$.
\end{proof}

\end{document}